\documentclass{article}

\usepackage[preprint]{corl_2026}
\usepackage[utf8]{inputenc}
\usepackage[T1]{fontenc}
\usepackage{hyperref}
\usepackage{url}
\usepackage{booktabs}
\usepackage{amsmath}
\usepackage{amsfonts}
\usepackage{nicefrac}
\usepackage{microtype}
\usepackage{xcolor}
\usepackage{array}
\usepackage{tabularx}

\usepackage{booktabs}
\usepackage[table]{xcolor}
\usepackage{float}
\usepackage{graphicx}
\graphicspath{{figures/}{./}}

\usepackage{wrapfig}

\title{Representation-Aligned Tactile Grounding for Contact-Rich Robotic Manipulation}

\author{
\textbf{Ruilin Chen$^{1}$}\thanks{Equal contribution.} \quad
\textbf{Jingkai Jia$^{2}$}\footnotemark[1] \quad
\textbf{Tong Yang$^{1,5}$} \quad
\textbf{Xinyu Zhou$^{4}$} \quad
\textbf{Qiao Sun$^{3}$} \quad
\textbf{Jiangwei Zhong $^{3}$} \\
\textbf{Shizeng Zhang$^{3}$} \quad
\textbf{Nuo Chen$^{3}$} \quad
\textbf{Bailin He$^{3}$} \quad
\textbf{Wei Li$^{2}$} \quad
\textbf{Wenqiang Zhang$^{1,2}$}\\
$^{1}$Shanghai Key Lab of Intelligent Information Processing, \\
College of Computer Science and Artificial Intelligence, Fudan University \\
$^{2}$College of Intelligent Robotics and Advanced Manufacturing, Fudan University\\
$^{3}$Lenovo CTO Organization \quad $^{4}$Nanyang Technological University \quad $^{5}$TeleAl, China Telecom\\
\texttt{\{rlchen25, jkjia24, tongyang23\}@m.fudan.edu.cn} \quad 
\texttt{sunqiao6@lenovo.com} \\
\texttt{wqzhang@fudan.edu.cn}
}

\begin{document}

\maketitle

\begin{abstract}
Tactile-enhanced vision-language-action (VLA) policies have been introduced for contact-rich manipulation, where critical interaction states are often hidden from vision. Future tactile prediction is a promising way to use touch because it turns tactile outcomes into supervision for action-induced contact dynamics. Yet VLA policies contain representations with different roles, from perceptual encoding to motor prediction, making it unclear where this supervision should be applied. We study this as a representation-alignment problem. Through a linear probe analysis, we find that future tactile states are most predictable from intermediate action-expert features, rather than from vision-language features or final action states. Motivated by this observation, we introduce a lightweight Latent Tactile Predictor (LTP), which predicts compact future tactile embeddings from the identified intermediate representation. By avoiding direct prediction of noisy raw tactile signals, LTP provides an action-outcome grounding signal that aligns intermediate action representations with future contact consequences. Experiments on real-world contact-rich manipulation tasks show that representation-aligned tactile grounding outperforms less aligned or multi-interface tactile prediction, highlighting the importance of where tactile supervision is applied.
\end{abstract}

\section{Introduction}

Despite recent progress in vision-language-action (VLA) models~\citep{kim2024openvla,black2024pi0,shukor2025smolvla}, contact-rich manipulation remains challenging because many critical state variables are not visually observable. Tasks such as insertion, alignment, and force-sensitive handling depend on robot--object--environment contact dynamics, where success hinges on how the robot presses, slides, and corrects its pose. Tactile sensing therefore provides a direct source of information about interaction dynamics hidden from visual perception~\citep{higuera2025sparshx,pattabiraman2024visk}, motivating tactile-enhanced VLA methods that introduce touch for contact-rich control~\citep{hao2025tla,huang2025tactilevla,zhang2026tacvla,li2026forcevla2}. Beyond conditioning the policy on tactile observations, future tactile prediction offers a more active way to use touch: it turns tactile outcomes into supervision for how the policy's own action chunk changes subsequent contact states, encouraging action-conditioned representations to encode contact dynamics.

\begin{figure}[t!]
\centering
\includegraphics[width=0.95\linewidth]{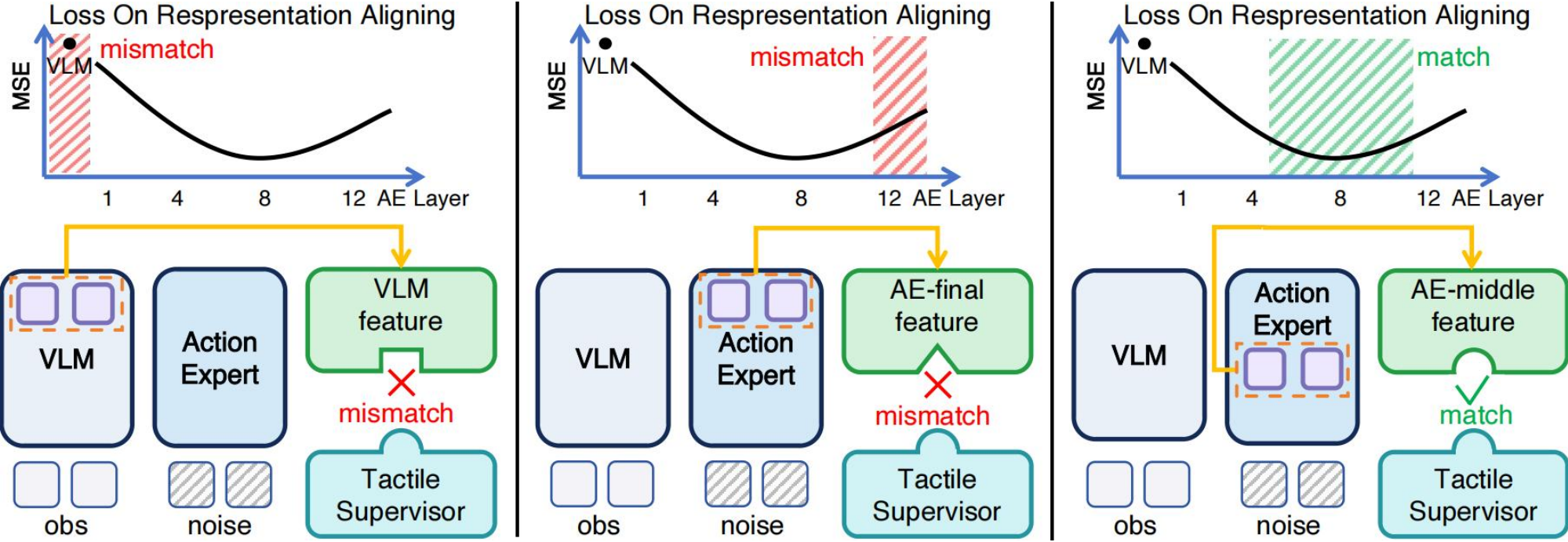}
\caption{\textbf{Motivation of representation-aligned tactile grounding.}
Future tactile prediction is most effective when applied to a representation aligned with action-conditioned contact dynamics. VLM-side features are too perception-oriented, while final action features are already specialized for motor decoding. We therefore ground the intermediate action-expert representation, where future tactile information is most accessible.}
\label{fig:motivation}
\end{figure}

Once tactile prediction is introduced, the key question is what representation it should shape. A natural intuition is to attach future tactile prediction to the final action representation, since tactile feedback and motor control are tightly coupled in contact-rich manipulation. Yet the final action state is already specialized for immediate motor decoding and may have compressed away information useful for modeling future contact. Conversely, applying tactile prediction to the vision-language backbone may keep the objective too close to semantic perception and too far from the action-conditioned dynamics that give tactile signals their control relevance. Thus, tactile prediction is not simply an auxiliary loss to be added anywhere in the policy. This raises a representation-alignment question: which internal representation should future tactile prediction supervise to ground action-conditioned contact dynamics rather than introduce mismatched regularization?

In this work, we study tactile prediction as an interaction-grounding objective for VLA policies. Rather than choosing the grounding interface by architectural convention, we diagnose where future tactile information is most accessible along the action pathway using lightweight frozen linear probes. Across settings with and without tactile input, as well as across model scales, the same pattern emerges: intermediate action-expert representations are more predictive of future tactile states than either vision-language backbone features or terminal action states. This suggests that future contact information is most available after action-conditioned interaction features have formed, but before the representation becomes specialized for immediate motor prediction.

Guided by this diagnosis, we apply tactile prediction to the intermediate action-expert representation. However, directly predicting raw tactile signals is ill-suited for grounding: tactile measurements are high-dimensional, sensor-specific, and noisy, so a raw prediction loss may emphasize sensing artifacts rather than control-relevant contact dynamics. We therefore introduce a lightweight Latent Tactile Predictor (LTP), which predicts future tactile outcomes in a compact latent space and provides a training-time grounding loss without changing the inference pathway. This loss encourages the intermediate action representation to preserve contact dynamics before they are compressed into immediate motor commands.

We evaluate this representation-aligned tactile grounding strategy across a range of real-world contact-rich manipulation tasks. Applying LTP at the intermediate action-expert representation achieves 74\% average success, outperforming less aligned tactile prediction interfaces at the VLM and final action states by 16 and 12 points, respectively, as well as broad multi-interface supervision. These results support the central premise of our method: tactile prediction benefits most when it supervises the stage where action-conditioned interaction dynamics are still being formed, before the final action layers specialize the representation toward executable motor predictions.

In summary, our contributions are as follows:
\begin{itemize}
    \item \textbf{Representation-aligned diagnosis.}
    We formulate future tactile prediction as representation-aligned grounding and use frozen linear probes to identify intermediate action-expert features as the most predictive interface for future tactile states.

    \item \textbf{Latent tactile grounding.}
    LTP predicts compact future tactile latents at the diagnosed intermediate interface, avoiding noisy raw tactile prediction while providing a training-time grounding loss without changing inference.

    \item \textbf{Effectiveness on contact-rich manipulation.}
    Our representation-aligned tactile grounding improves  contact-rich manipulation and outperforms less aligned and multi-interface tactile prediction strategies.
\end{itemize}


\section{Related Work}
\label{sec:related_work}

\subsection{Vision-Language-Action Models}
By leveraging the semantic and perceptual capabilities of pre-trained
vision-language models~\cite{liu2023visual,karamcheti2024prismatic,
marafioti2025smolvlm}, vision-language-action (VLA) models extend
multimodal pretraining to robotic control by mapping visual observations
and language instructions into executable actions~\cite{brohan2022rt,
brohan2023rt,team2024octo,kim2024openvla,black2024pi0,
intelligence2025pi_,shukor2025smolvla}. These models differ in how they
adapt pretrained backbones and parameterize robot actions, but they share
the same central objective of translating visual-linguistic context into
executable action trajectories. This line of work has substantially improved
the scalability, generality, and deployability of robot policies. However, these advances mainly improve how actions are generated from visual-linguistic context, while contact states remain difficult to observe. States such as slip, pressure, local resistance, and insertion alignment are often visually ambiguous or hidden, but are decisive for contact-rich manipulation. This contact-observability gap motivates the use of tactile sensing, which provides direct access to interaction states that vision alone may miss~\citep{pattabiraman2024visk,higuera2025sparshx,zhu2025touchwild}. It has also motivated recent efforts to extend VLA policies with tactile feedback.

\subsection{Tactile-Enhanced Vision-Language-Action Models}

Motivated by this contact-observability gap, tactile-enhanced VLA methods provide policies with direct access to touch. Existing work mainly improves how tactile signals enter or modulate the policy, through tactile tokens or embeddings, multimodal fusion, adapters, gating, or mixture-of-experts mechanisms~\citep{hao2025tla,zhang2025vtla,huang2025tactilevla,zhang2026tacvla,cheng2025omnivtla,bi2025vlatouch,yu2025forcevla,li2026forcevla2,morissette2026tacfilm,gubernatorov2026hapticvla,huang2026tafvla}. These designs improve tactile access and fusion, but mostly treat touch as an input signal: the policy can condition on tactile readings, yet is not explicitly required to model how its own actions change future contact.

A complementary direction is to predict future tactile states or visuo-tactile dynamics conditioned on actions~\citep{liu2025mla,xu2025exumi,higuera2026vtwm,ye2025dreamtacvla}. Prediction turns tactile outcomes into supervision for action-induced contact dynamics, making touch more active than passive input. However, existing work typically uses such prediction as an auxiliary loss, world-modeling objective, or feedback signal, without studying which internal policy representation should be shaped by this supervision. Our work addresses this representation-alignment problem by diagnosing where future tactile prediction should supervise the VLA action pathway and using it to design a representation-aligned tactile grounding objective.

\section{Method}

\subsection{Overview}
\label{sec:method:overview}

Future tactile prediction can ground VLA policies in contact dynamics, but only if the loss is applied to a suitable representation. Since future tactile targets describe contact consequences of the policy’s action chunk, they are best suited to supervise representations that are action-conditioned but not yet specialized for motor prediction. This creates a representation-alignment problem: VLM features may be too semantic, while final action states may be too control-specialized.
To resolve this alignment problem, we first probe the frozen policy to identify where future tactile information is most accessible along the action pathway. This diagnosis reveals intermediate action-expert features as the most suitable grounding interface. We then attach a lightweight Latent Tactile Predictor (LTP) at this interface, predicting compact future tactile latents to avoid raw sensor noise.

\begin{figure}[t]
\centering
\includegraphics[width=1\linewidth]{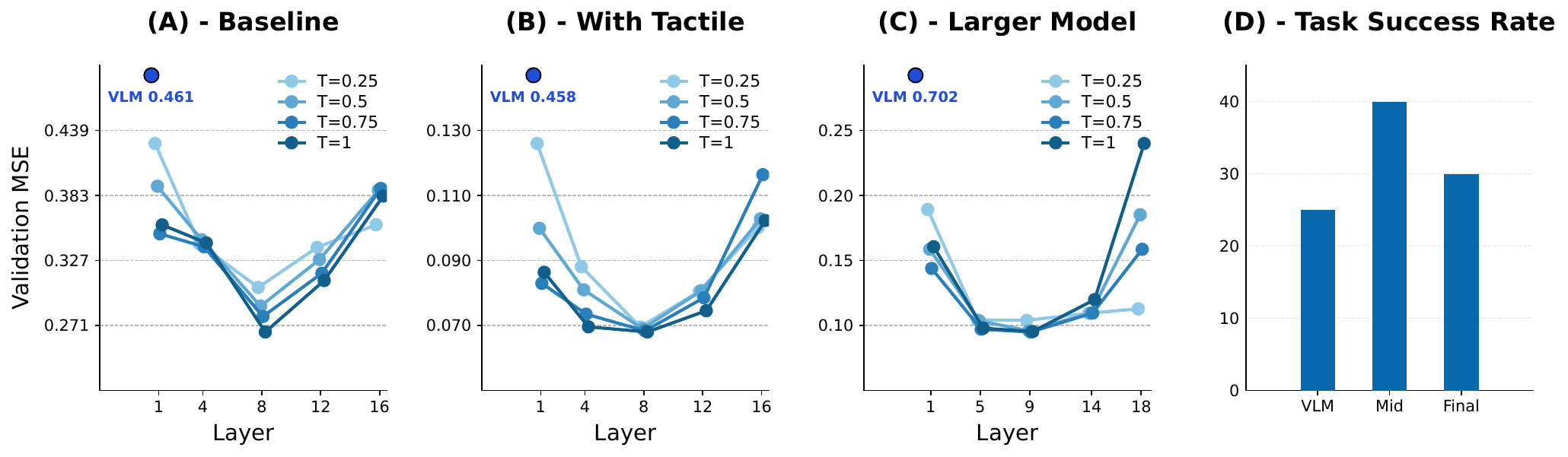}
\caption{\textbf{Linear-probe diagnosis and downstream validation.} (A) Future tactile prediction MSE without tactile input. (B) The same probe analysis with current tactile as input. (C) The same analysis on the $\pi_0$ backbone. (D) Downstream success when grounding VLM-side, intermediate
action-expert, and final action representations. Curves in (A--C) are evaluated at normalized action-generation timesteps \(T\) of the flow-based action expert.
}
\label{fig:linear-prob-diag}
\end{figure}

\subsection{Diagnosing the Tactile Grounding Interface}
\label{sec:method:prediction_grounding}

Before introducing the tactile grounding loss, we diagnose where future tactile information is most accessible along the VLA action pathway. We start from a base VLA policy trained with its native action objective and freeze all policy parameters. For a trajectory segment starting at time $t$, we extract candidate representations from the VLM-side output $h_t^{\mathrm{vlm}}$ and each action-expert layer $\{h_t^\ell\}_{\ell=1}^{L}$. For each candidate representation $h$, we train an independent lightweight linear probe $g_h$ to predict the raw future tactile sequence $r_{t+1:t+H}$:
\begin{equation}
\hat{r}_{t+1:t+H} = g_h(h), \qquad
\mathcal{L}_{\mathrm{probe}}
=
\left\|
\hat{r}_{t+1:t+H}
-
r_{t+1:t+H}
\right\|_2^2 .
\end{equation}
The policy is never updated during this diagnosis. Lower probe error indicates that future tactile information is more linearly accessible from the corresponding representation. We use raw future tactile measurements as the probe target to keep the interface diagnosis independent of the latent tactile encoder used later for policy training. For flow-based action experts, we repeat the probe at several normalized action-generation times $T \in \{0.25,0.5,0.75,1.0\}$ and report the predictability trend across candidate representations for each $T$.

\paragraph{Future tactile information peaks at intermediate action-expert layers.}
We first examine the clean diagnostic setting, where the SmolVLA policy is frozen and tactile observations are removed from the input. As shown in Fig.~\ref{fig:linear-prob-diag}(A), future tactile predictability follows a non-monotonic trend along the action pathway. The VLM-side representation has high prediction error, indicating that future contact outcomes are not easily read out from perceptual-semantic features alone. Prediction error decreases as representations enter the action expert and reaches its minimum at an intermediate layer, before increasing again near the final action state.

This pattern suggests that future tactile information is most accessible after the policy has formed action-conditioned interaction features, but before these features are compressed toward immediate motor prediction. The VLM-side representation remains close to semantic perception, while the final action state is specialized for executable action decoding. The intermediate action-expert representation lies between these extremes and therefore provides a better-matched interface for future tactile supervision. We observe the same qualitative trend when tactile observations are included as input and when the diagnosis is repeated with a larger VLA backbone, as shown in Fig.~\ref{fig:linear-prob-diag}(B,C). This indicates that the intermediate-stage advantage is not an artifact of removing tactile input or of a specific model scale.

\paragraph{Tactile predictability aligns with grounding effectiveness.}
We next test whether the representation with the strongest tactile predictability also provides the best grounding interface for policy learning. We apply future tactile prediction to representative interfaces and evaluate the resulting policies on the USB Drive Insertion task. As shown in Fig.~\ref{fig:linear-prob-diag}(D), downstream success across these evaluated interfaces follows the same ordering as the probe results: VLM-side grounding is weakest, final-state grounding improves but remains suboptimal, and intermediate action-expert grounding performs best. This agreement supports using the probe-diagnosed intermediate representation as the interface for latent tactile grounding.

\subsection{Latent Tactile Predictor}
\label{sec:method:latent_tactile_prediction}
\begin{figure}[t!]
\centering
\includegraphics[width=0.95\linewidth]{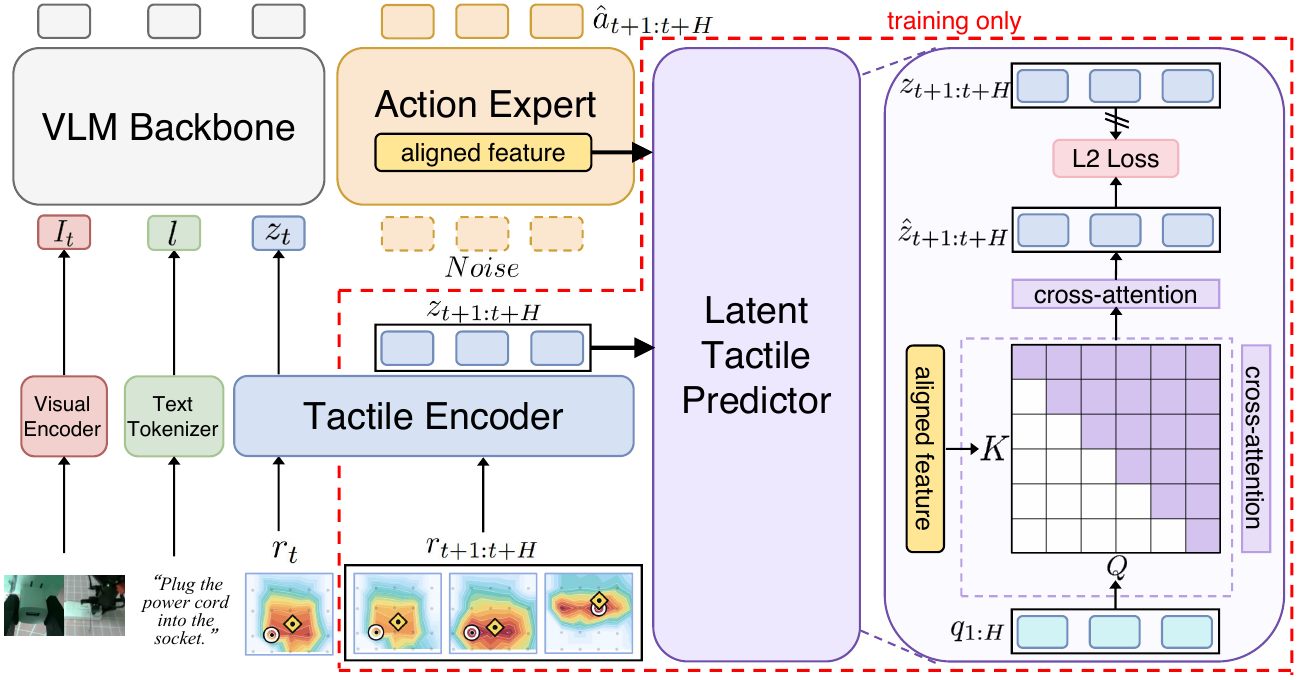}
\caption{
\textbf{Overview of future latent tactile grounding.} The VLA policy takes current observations $(I_t,l,r_t)$  - image, language, and tactile input—and predicts an action chunk $\hat{a}_{t+1:t+H}$. During training, the aligned action-expert feature, which is most grounded in future contact states,  conditions a latent tactile predictor, which predicts future tactile latents $\hat{z}_{t+1:t+H}$ from learnable queries $q_{1:H}$. The targets $z_{t+1:t+H}$ are encoded from future tactile observations $r_{t+1:t+H}$ and used with stop-gradient for the grounding loss. The predictor is used only during training.
}
\label{fig:method-plot}
\end{figure}
Having identified the intermediate action-expert representation as the grounding interface, we now define the latent tactile prediction objective applied to this representation. As illustrated in Fig.\ref{fig:method-plot} , at time $t$, the tactile-conditioned VLA policy receives the current image observation $I_t$, language instruction $l$, and current tactile observation $r_t$. The current tactile observation $r_t$ is encoded into tactile latent tokens and provided to the policy together with the visual-language and state tokens. The policy then predicts an action chunk $\hat{a}_{t+1:t+H}$ through its native action-generation pathway. After executing this action chunk, the robot observes a future tactile sequence $r_{t+1:t+H}$, which is used only during training as the tactile grounding target. Thus, current tactile $r_t$ serves as policy input, while future tactile $r_{t+1:t+H}$ serves as supervision.

Directly predicting the raw future tactile sequence $r_{t+1:t+H}$ is ill-suited for grounding. Tactile signals are high-dimensional, sensor-dependent, and often contain measurement noise, calibration bias, and local contact artifacts. A raw tactile prediction objective may therefore encourage the policy representation to model sensor recovery rather than the contact consequences useful for action generation. We instead predict future tactile outcomes in a compact latent space, so that the auxiliary objective focuses on contact-relevant structure rather than raw sensor details.

Concretely, we encode the future tactile sequence into a latent tactile
sequence:
\begin{equation}
z_{t+1:t+H}
=
E_{\mathrm{tac}}(r_{t+1:t+H}),
\end{equation}
where $E_{\mathrm{tac}}$ denotes the tactile encoder in our framework. Let $h_t^{\mathrm{mid}}$ be the intermediate action-expert representation identified by the probe analysis. The Latent Tactile Predictor uses a set of learnable tactile queries $Q=\{q_1,\ldots,q_H\}$ to predict the future tactile latent sequence from the intermediate action-expert representation:
\begin{equation}
\hat{z}_{t+1:t+H}
=
P_{\theta}\left(Q, h^{\mathrm{mid}}_t\right),
\end{equation}
where $P_{\theta}$ is a lightweight query-based prediction head.

We train the predictor with a latent tactile prediction loss:
\begin{equation}
\mathcal{L}_{\mathrm{tac}}
=
\left\|
\hat{z}_{t+1:t+H}
-
\mathrm{sg}(z_{t+1:t+H})
\right\|_2^2,
\end{equation}
where $\mathrm{sg}(\cdot)$ stops gradients through the tactile target. The final training objective combines this grounding loss with the native VLA
action objective:
\begin{equation}
\mathcal{L}
=
\mathcal{L}_{\mathrm{act}}
+
\lambda_{\mathrm{tac}}
\mathcal{L}_{\mathrm{tac}} .
\end{equation}

The LTP branch is used only during training. At inference, the prediction branch is removed, and the policy keeps the original tactile-conditioned action-generation pathway unchanged. In this way, future tactile prediction provides a compact grounding signal for the intermediate action representation without adding inference-time cost.

\section{Experiments}
\label{sec:experiments}
\begin{figure}[t]
\centering
\includegraphics[width=0.95\linewidth]{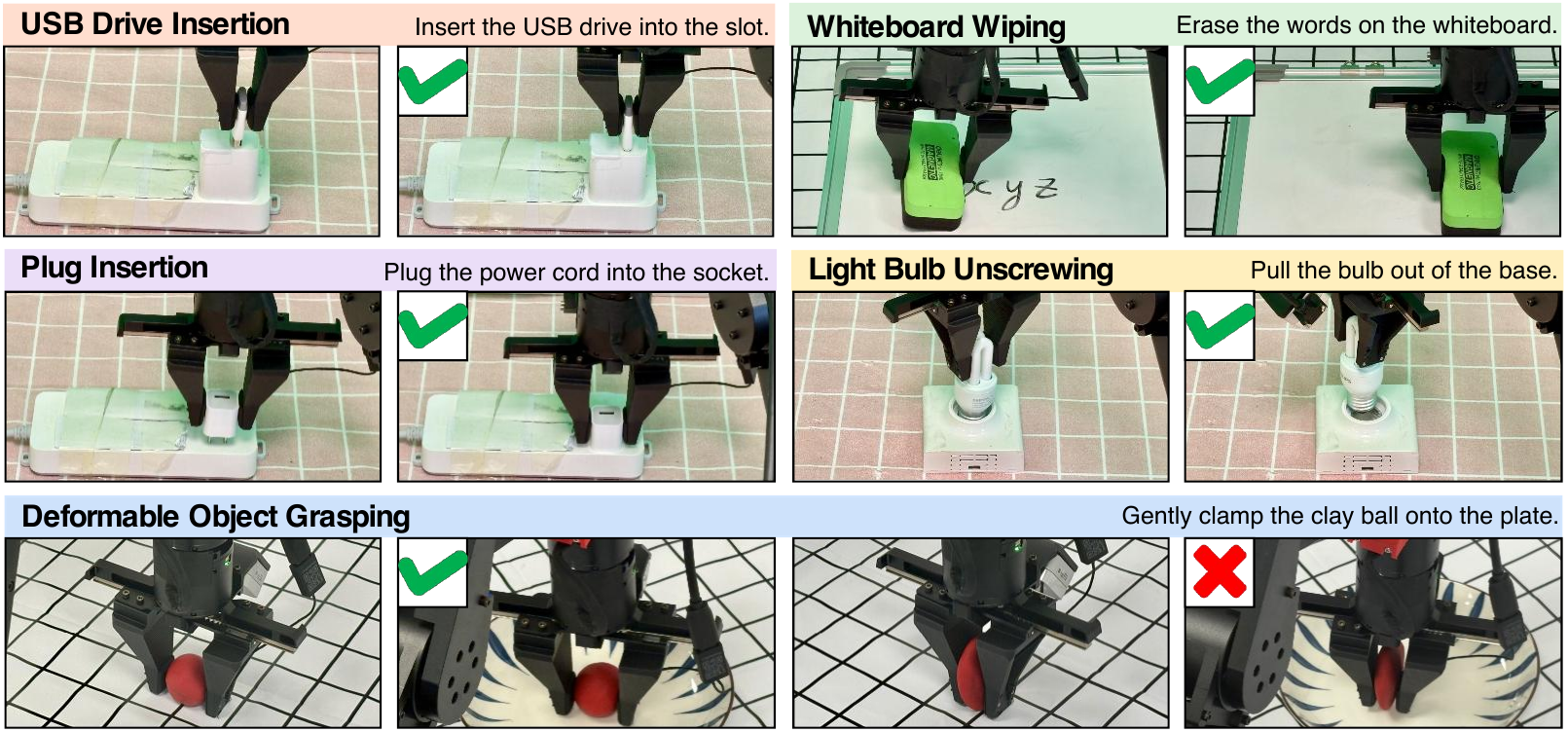}
\caption{
\textbf{Experimental task suite.}
The five blocks correspond to USB Drive Insertion, Whiteboard Wiping, Plug Insertion, Light Bulb Unscrewing and Deformable Object Grasping.Green check marks indicate successful executions. For deformable-object grasping, both a successful execution and a failure case, marked with a red cross, are shown.
}
\label{fig:exp-setup}
\end{figure}

We evaluate whether representation-aligned tactile grounding enables VLA policies to use touch more effectively in real-world contact-rich manipulation. Rather than treating tactile sensing as a binary design choice, our experiments focus on how future tactile supervision should be used once tactile information is available to the policy. We therefore compare different grounding interfaces, tactile prediction targets, and supervision scopes under the same tactile-conditioned VLA setting.

The experiments are designed to answer the following questions:
\begin{itemize}
\item Does the diagnosed intermediate action-expert interface outperform less aligned tactile prediction interfaces?

\item Does predicting compact future tactile latents provide a stronger grounding signal than raw tactile prediction?

\item Is representation-aligned single-interface grounding more effective than indiscriminate multi-interface tactile supervision?
\end{itemize}

\subsection{Experimental Setup}
\label{sec:exp_setup}

\paragraph{Tasks and data collection.}
All experiments are conducted on a real-world single-arm setup with an ARX R5 robot equipped with PaXini tactile sensors. The policy observes two RGB streams, including a third-person camera and a wrist-mounted camera, together with robot proprioception and tactile readings. We evaluate on five contact-rich manipulation tasks designed to cover complementary physical interaction challenges. For each task, we collect 50 expert demonstrations using leader--follower teleoperation, recording synchronized RGB observations, robot states, actions, and tactile readings throughout execution.

Fig. 4 illustrates the task suite, while the text below summarizes the contact-specific capability tested by each task.
\textit{Plug Insertion} and \textit{USB Drive Insertion} test fine alignment
and force-sensitive insertion under tight geometric tolerance. \textit{Whiteboard
Wiping} tests stable surface contact and pressure regulation during continuous
motion: insufficient pressure fails to erase the mark, while excessive pressure
can push the board away. \textit{Deformable Object Grasping} uses a plasticine
object to evaluate pressure-sensitive grasping without visible deformation.
\textit{Light Bulb Unscrewing} tests contact engagement, alignment maintenance,
and controlled rotational motion.


\paragraph{Baselines.}
We organize baselines around the alternative explanations they test. Standard VLA policies (SmolVLA~\citep{shukor2025smolvla} and $\pi_0$~\citep{black2024pi0}) measure performance without tactile grounding. Tactile-conditioned variants receive the same tactile stream as input, testing whether direct tactile access is sufficient. exUMI~\citep{xu2025exumi} provides a predictive tactile representation baseline, where future tactile prediction improves the tactile encoder before policy learning.
In contrast, our method uses future tactile outcomes to supervise an internal action representation. To isolate this representation-alignment effect, we also evaluate interface-controlled variants that attach the same tactile prediction loss to the vision-language backbone output, the final action state, or the diagnosed intermediate action-expert feature. Finally, multi-interface grounding variants add tactile losses to multiple layers, testing whether performance comes from alignment rather than loss accumulation.

\begin{figure}[t]
\centering
\includegraphics[width=0.95\linewidth]{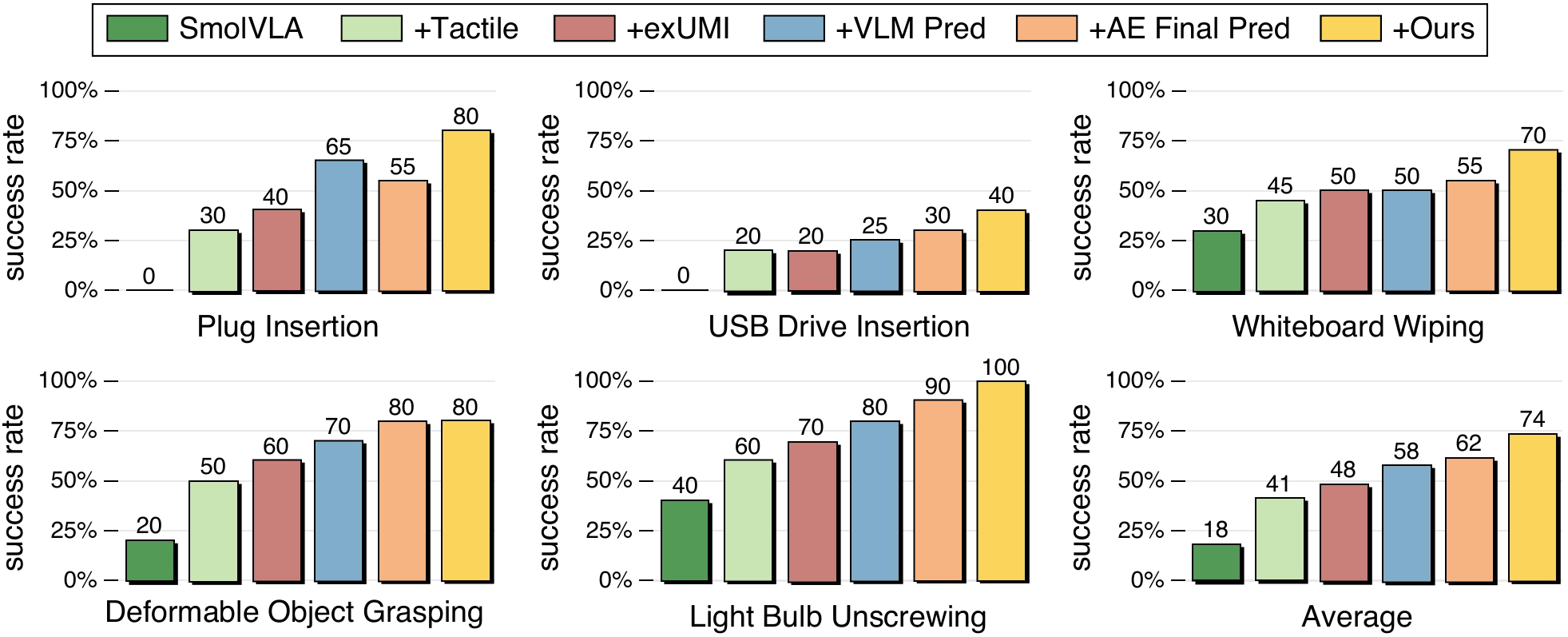}
\caption{
\textbf{Success rates on five real-world contact-rich manipulation tasks using the SmolVLA backbone.} We report task success rates for input-level tactile conditioning, external predictive tactile representation learning, less-aligned tactile prediction interfaces, and our representation-aligned tactile grounding under identical settings.
}
\label{fig:smolvla-main-result}
\end{figure}

\paragraph{Training protocol.}
Unless otherwise specified, all methods are trained with the same demonstrations, observation streams, tactile inputs when applicable, and action representation. For SmolVLA-based methods, we use a batch size of 64, train for 200K steps, and set the learning rate to $1\times10^{-4}$. For $\pi_0$-based methods, we follow the standard LoRA fine-tuning protocol under the same data setting. Baseline implementations follow their default training configurations when available.

\paragraph{Evaluation protocol.}
We evaluate each method with 20 real-world trials per task and report task success rate as the main metric. Success is judged by task-specific criteria, including correct insertion, sufficient wiping coverage with stable contact, lifting without visible deformation, and successful unscrew engagement and tightening.

\subsection{Main Results}
\label{sec:main_results}

\paragraph{Representation-aligned tactile grounding improves contact-rich manipulation.}
Fig.~\ref{fig:smolvla-main-result} compares different uses of tactile information in SmolVLA policies. Input-level tactile conditioning and external tactile representation learning improve over the vision-language-action baseline, but remain below representation-aligned grounding. The key comparison is among future tactile prediction interfaces: VLM-level prediction and final-action-state prediction achieve 58\% and 62\% average success, whereas grounding the intermediate action-expert representation reaches 74\%. This ordering indicates that future tactile prediction is not effective merely because it adds an auxiliary objective; its benefit depends on applying the supervision to a representation aligned with action-conditioned contact dynamics.

The task-wise results show that the improvement is not driven by a single task: our method is best on four tasks and ties for best on Deformable Object Grasping. Plug Insertion is especially informative: VLM-level prediction outperforms final-action-state prediction, while intermediate grounding remains best, suggesting that tactile prediction does not simply improve by moving closer to the motor output. USB Drive Insertion remains the hardest task, but our method still achieves the highest success rate. These patterns support the representation-alignment interpretation behind the aggregate improvement.

\begin{figure}[t!]
\centering
\includegraphics[width=0.95\linewidth]{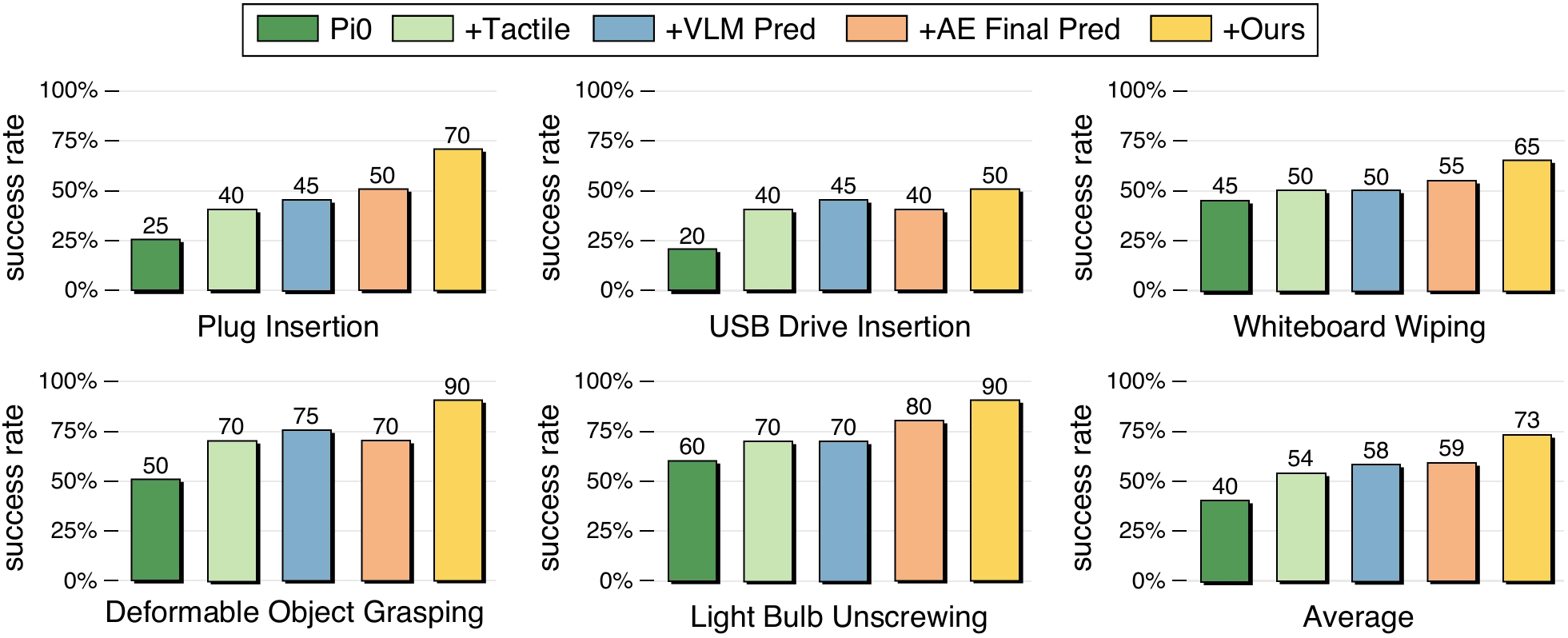}
\caption{\textbf{Success rates with the $\pi_0$ backbone on five real-world contact-rich manipulation tasks.} We compare the standard $\pi_0$ policy, tactile-conditioned $\pi_0$, two less-aligned future tactile prediction interfaces (VLM Pred. and AE Final Pred.), and our representation-aligned tactile grounding under the same task setting.}
\label{fig:pi0-main-results}
\end{figure}

\paragraph{Generalization to a different VLA backbone.}
We further evaluate our grounding objective with the $\pi_0$ backbone under LoRA fine-tuning. Since this adaptation protocol differs from the SmolVLA setting, we focus on relative gains within the $\pi_0$ family rather than absolute comparison across backbones. As shown in Fig.~\ref{fig:pi0-main-results}, we additionally include two interface-controlled prediction baselines, which attach the same future tactile prediction objective to the VLM-side representation or the final action-expert state. Under the same contact-rich manipulation setting, the standard $\pi_0$ policy achieves 40\% average success, and tactile conditioning improves it to 54\% within this controlled evaluation. VLM-side and final-action prediction further reach 58\% and 59\%, respectively, but still remain below our representation-aligned grounding, which achieves 73\% average success with clear gains across the task suite. This suggests that the diagnosed intermediate action-expert interface is not an artifact of a single backbone: future tactile prediction is most effective when applied to an action-conditioned representation aligned with contact dynamics, rather than simply added to the policy or attached to less aligned interfaces.

\subsection{Ablation Studies}
\label{sec:ablations}

\begin{figure}[t!]
\centering
\small
\begin{tabular*}{0.95\linewidth}{@{\extracolsep{\fill}}lcccccc@{}}
\toprule
\textbf{Method} & \textbf{Plug} & \textbf{USB} & \textbf{Wipe} & \textbf{Deformable} & \textbf{Bulb} & \textbf{Avg.} \\
\midrule
None & 30 & 20 & 45 & 50 & 60 & 41 \\
Multi-A & 20 & 20 & 45 & 45 & 60 & 38 \\
Multi-B & 30 & 25 & 55 & 55 & 70 & 48 \\
Multi-C & 50 & 30 & 60 & 70 & 90 & 60 \\
Ours & 80 & 40 & 70 & 80 & 100 & 74 \\
\bottomrule
\end{tabular*}
\caption{\textbf{Ablation results on multi-layer grounding applied to SmolVLA across all tasks.} None adds tactile input without future tactile prediction. Multi-A predicts from all action-expert features, Multi-B predicts from every two action-expert layers (0/2/4/...), and Multi-C predicts from the middle layers 4--11 of the 16-layer SmolVLA action expert.}
\label{fig:multi-layer-ablation}
\end{figure}

\paragraph{Latent Prediction.}
We first ablate the form of the future tactile target. Replacing compact tactile latents with direct raw tactile prediction reduces Plug Insertion success from 80\% to 55\%.
{\setlength{\columnsep}{3pt}%
\begin{wraptable}[4]{r}{0.305\linewidth}
\vspace{-9.5pt}
\centering
\setlength{\tabcolsep}{4.0pt}
\renewcommand{\arraystretch}{0.88}
\begin{tabular}{@{}l@{\hspace{0.75em}}c@{}}
\toprule
\textbf{Target} & \textbf{Success Rate} \\
\midrule
Raw & 55 \\
Ours & 80 \\
\bottomrule
\end{tabular}
\vspace{-10pt}
\end{wraptable}%
This result suggests that predicting future tactile signals in a compact latent space provides a stronger grounding signal than predicting raw tactile observations. Direct raw prediction can force the grounding loss to account for sensor-level details, which may include measurement noise, calibration bias, and local contact artifacts. Such low-level variations can provide a less suitable constraint for shaping action-conditioned representations. In contrast, the encoder-produced tactile latents offer a compact future-contact target, making the grounding objective more focused on action-relevant contact outcomes rather than raw signal recovery.
\par}

\paragraph{Effect of supervision scope.}
We next test whether future tactile prediction can simply be applied broadly across the action pathway, instead of using a diagnosed grounding interface. A natural alternative is to apply the same future tactile
prediction loss to multiple points in the action pathway. As shown in
Fig.~\ref{fig:multi-layer-ablation}, we extend this ablation to all five tasks.
SmolVLA with tactile input but without future tactile prediction (None) achieves
41\% average success. Applying the loss more broadly is consistently
suboptimal: grounding all candidate representations (Multi-A), every two layers
(Multi-B), or multiple middle representations (Multi-C) reaches 38\%, 48\%,
and 60\% average success, respectively, while our single aligned interface
achieves 74\%. These results show that the gain is not explained by broadly adding more tactile prediction losses. Applying the same contact-dynamic target to representations with different roles can introduce mismatched supervision, whereas the diagnosed interface provides a cleaner grounding signal for action generation across the task suite.

\section{Conclusion}
We presented representation-aligned tactile grounding for contact-rich robotic manipulation. Rather than treating future tactile prediction as a generic auxiliary loss, we showed that its effectiveness depends on which policy representation it supervises. Frozen linear probes identify intermediate action-expert representations as the most tactile-predictive interface, and our Latent Tactile Predictor uses compact future tactile latents to ground this representation during training without changing inference. Real-world experiments on five contact-rich manipulation tasks show consistent gains across SmolVLA and $\pi_0$, outperforming less aligned interfaces, raw tactile prediction, and multi-interface grounding. Overall, our results indicate that future tactile prediction benefits VLA policies most when used as representation-aligned grounding, rather than as tactile supervision alone.

\clearpage

\bibliography{reference}

\clearpage

\section*{Appendix}

\subsection*{A\quad Implementation Details}

For clarity, we summarize the training protocols used in our experiments in
Tables~\ref{tab:app_smolvla_hyperparameters} and~\ref{tab:app_pi0_lora_hyperparameters}.
We train tactile-supervised policies on two architectures, SmolVLA and $\pi_0$,
using datasets stored in the LeRobot format. Unless otherwise noted, both
settings use a latent action dimension of 7 and a tactile supervision weight of
1. The remaining architecture-specific choices, optimizer settings, learning
rate schedules, batch sizes, and training durations are reported in the
corresponding tables.

\begin{table}[H]
\centering
\caption{Training hyperparameters for SmolVLA.}
\label{tab:app_smolvla_hyperparameters}
\large
\begin{tabular}{ll}
\toprule
\textbf{Parameter} & \textbf{Value} \\
\midrule
Latent action dim & 7 \\
Tactile supervision weight $\lambda$ & 1 \\
\midrule
Training steps & 200k \\
Batch size & 64 \\
Chunk size & 16 \\
Cosine learning rate schedule & $1.0{\times}10^{-4} \rightarrow 2.5{\times}10^{-6}$ \\
Optimizer & AdamW \\
Weight decay & $1.0{\times}10^{-10}$\\
\bottomrule
\end{tabular}
\end{table}

\begin{table}[H]
\centering
\caption{Training hyperparameters for $\pi_0$ LoRA.}
\label{tab:app_pi0_lora_hyperparameters}
\large
\begin{tabular}{ll}
\toprule
\textbf{Parameter} & \textbf{Value} \\
\midrule
Latent action dim & 7 \\
Tactile supervision weight $\lambda$ & 1 \\
\midrule
Training steps & 30k \\
Batch size & 64 \\
Chunk size & 16 \\
Cosine learning rate schedule & $1.0{\times}10^{-4} \rightarrow 1.0{\times}10^{-5}$ \\
Optimizer & AdamW \\
Weight decay & $0$ \\
LoRA  rank & 16 \\
\bottomrule
\end{tabular}
\end{table}

\clearpage

\subsection*{B\quad Qualitative Results of Real-world Experiments}

\newcommand{\appframeimg}[1]{%
  \includegraphics[width=0.1666\linewidth]{#1}%
}

\newcommand{\apptaskrow}[8]{%
  \appframeimg{#2}%
  \appframeimg{#3}%
  \appframeimg{#4}%
  \appframeimg{#5}%
  \appframeimg{#6}%
  \appframeimg{#7}

  \vspace{0.08cm}
  {\large\textbf{#1} #8}
}

\begin{figure}[H]
\centering
\apptaskrow{(a)}{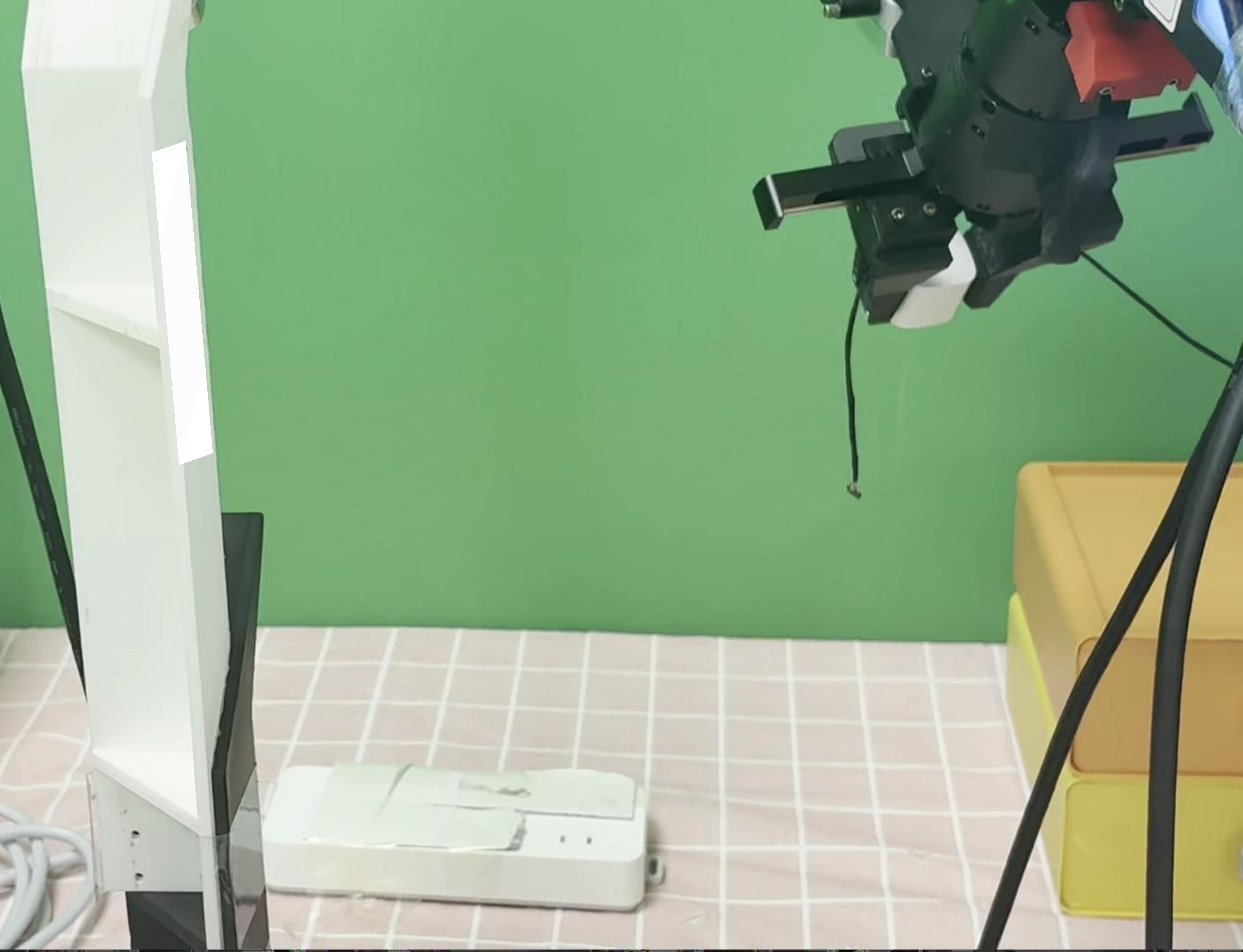}{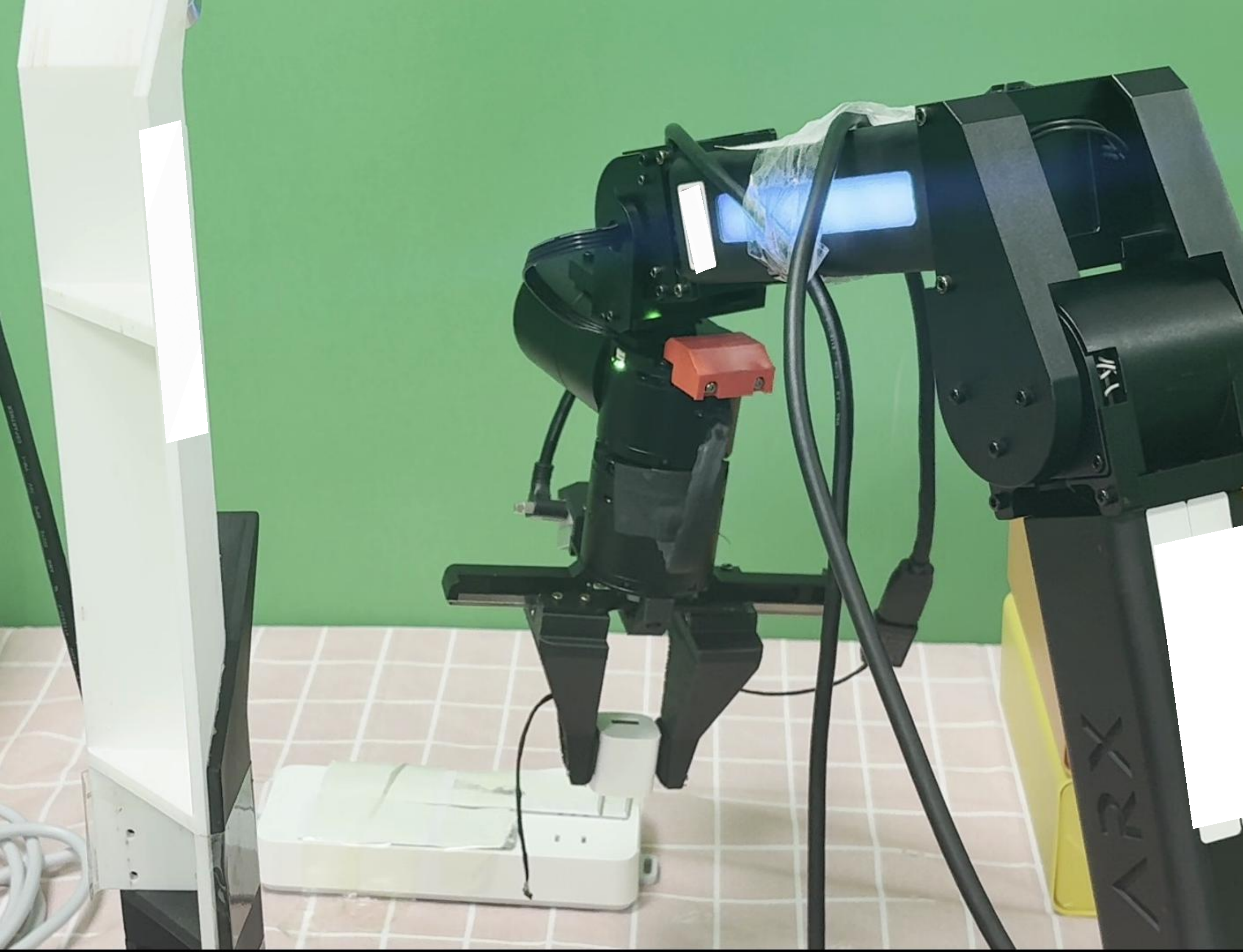}{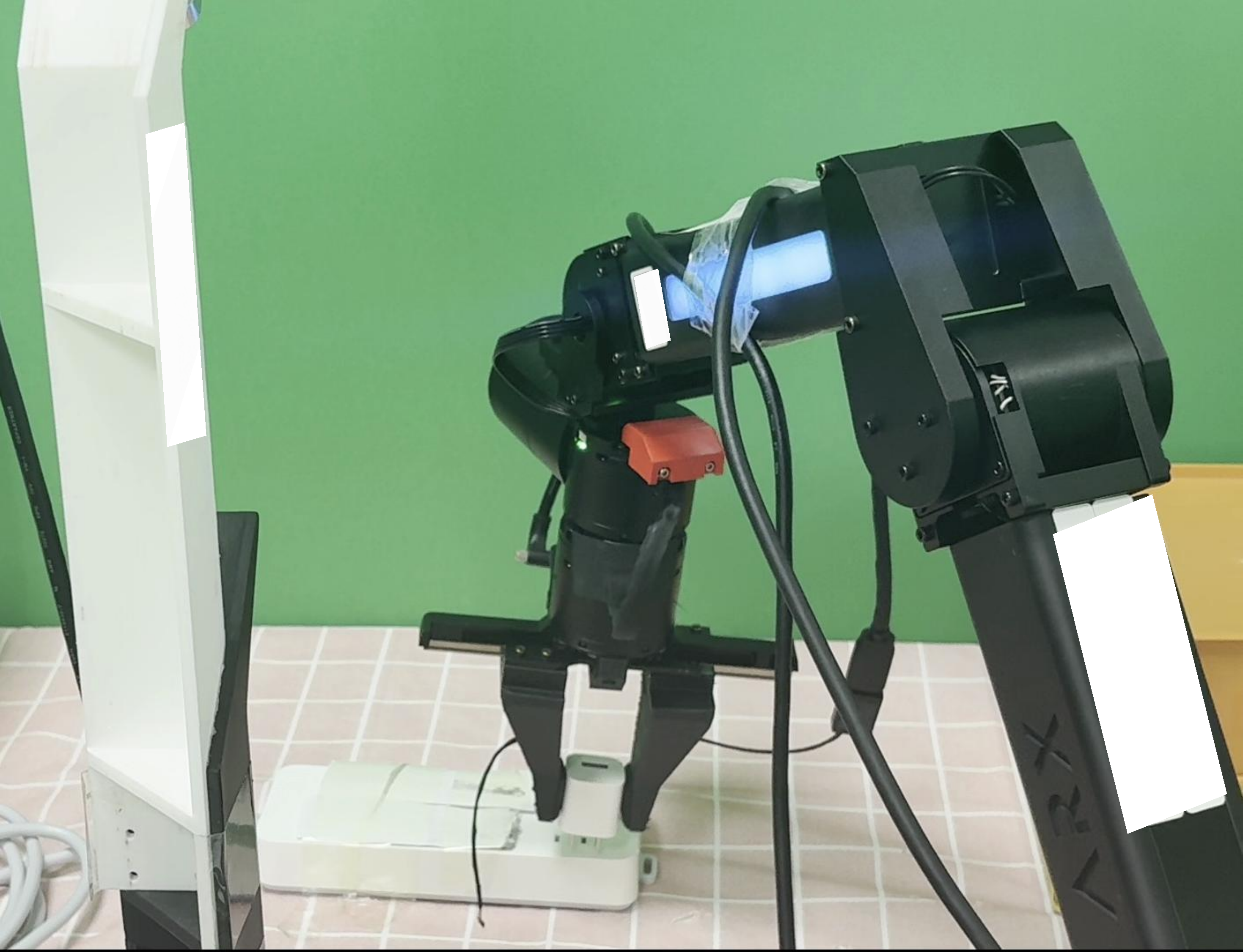}{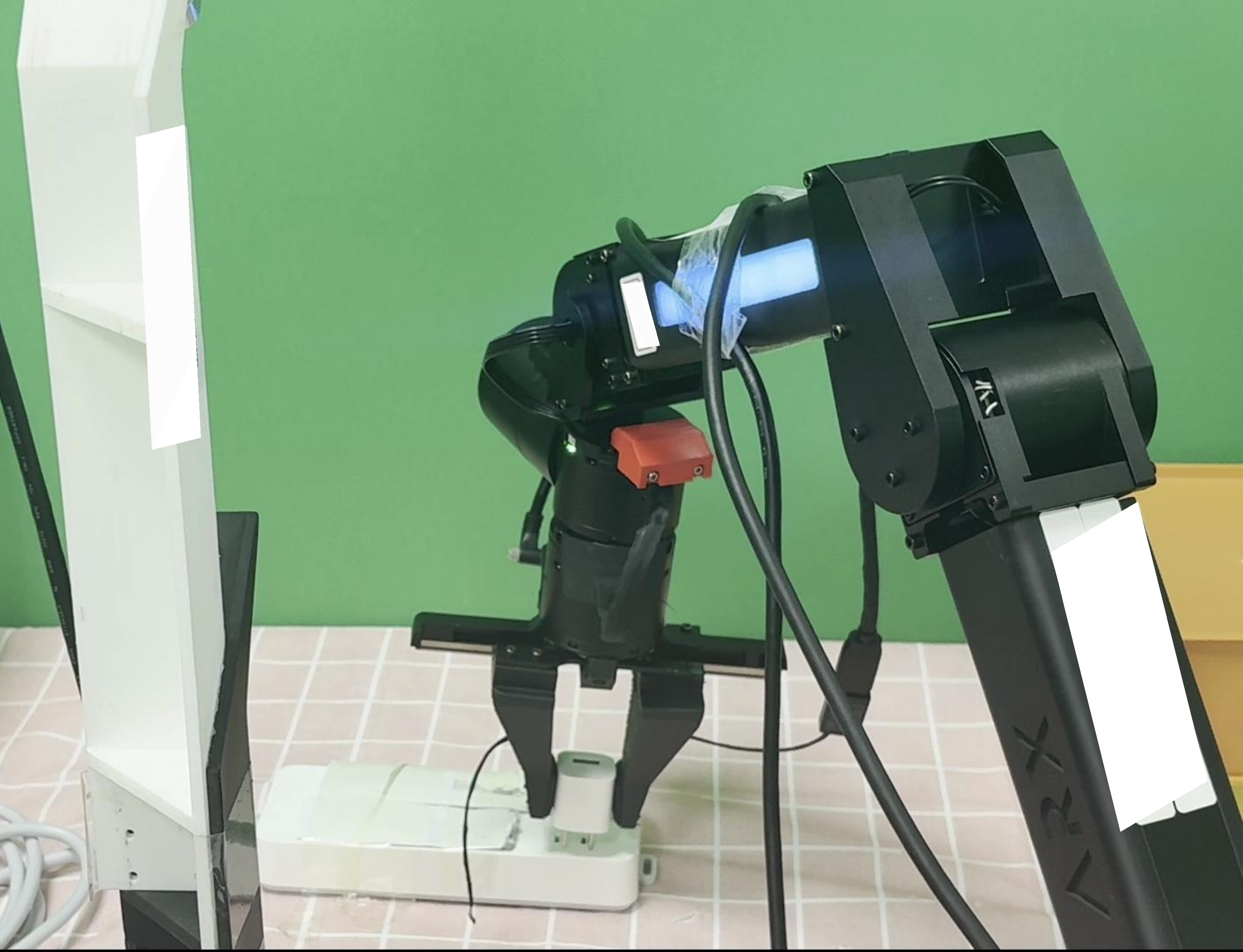}{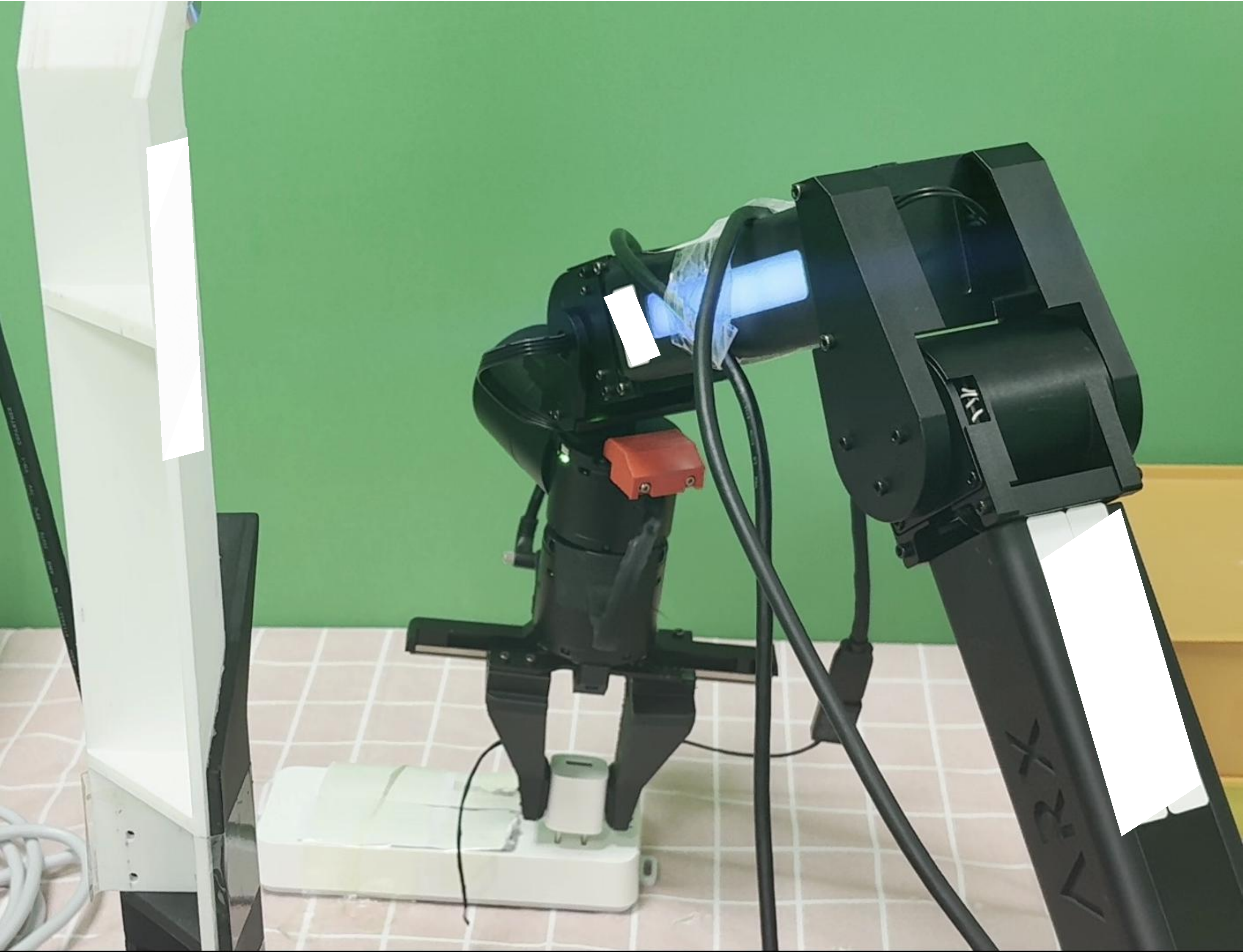}{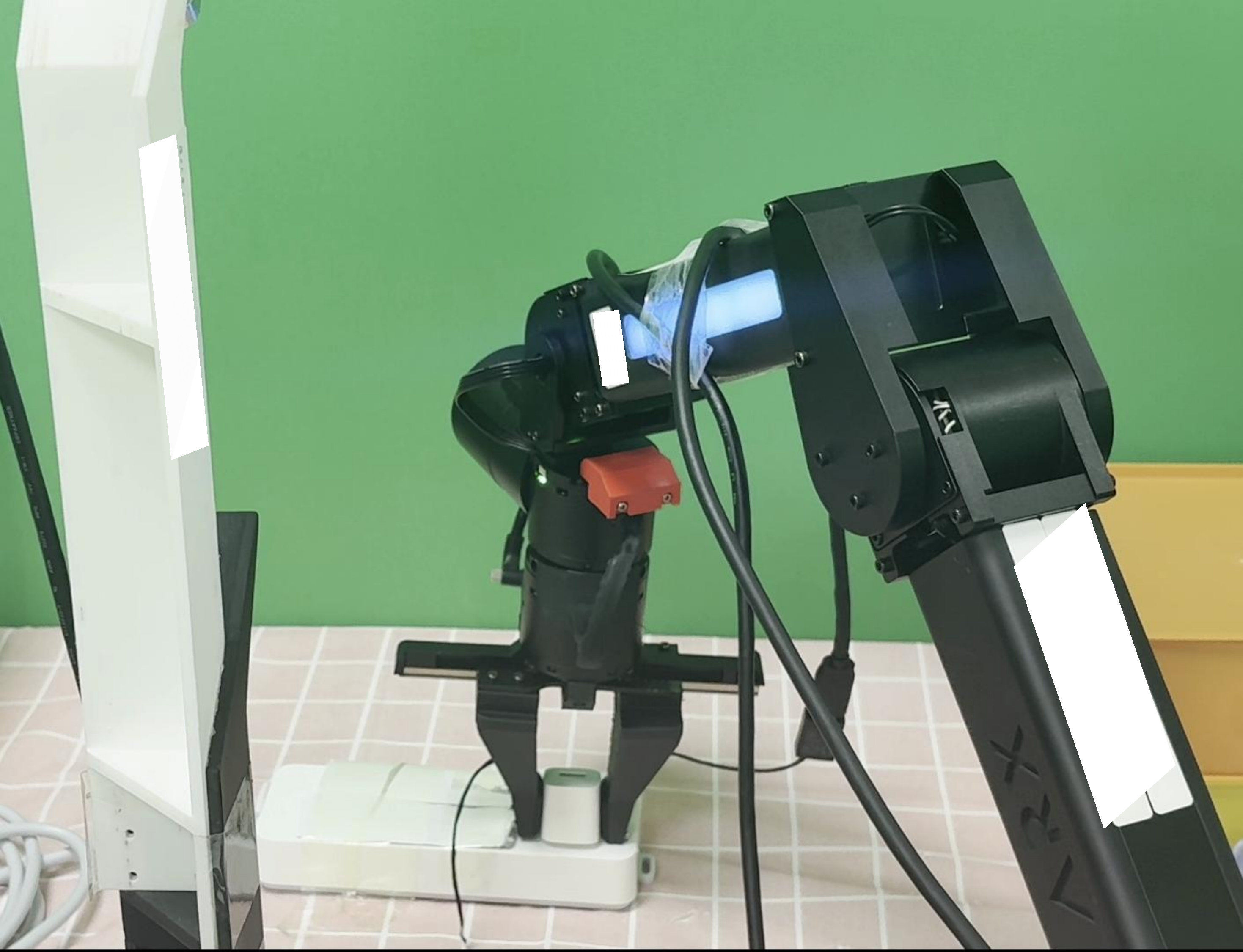}{Plug Insertion : Plug the power cord into the socket.}

\vspace{0.24cm}
\apptaskrow{(b)}{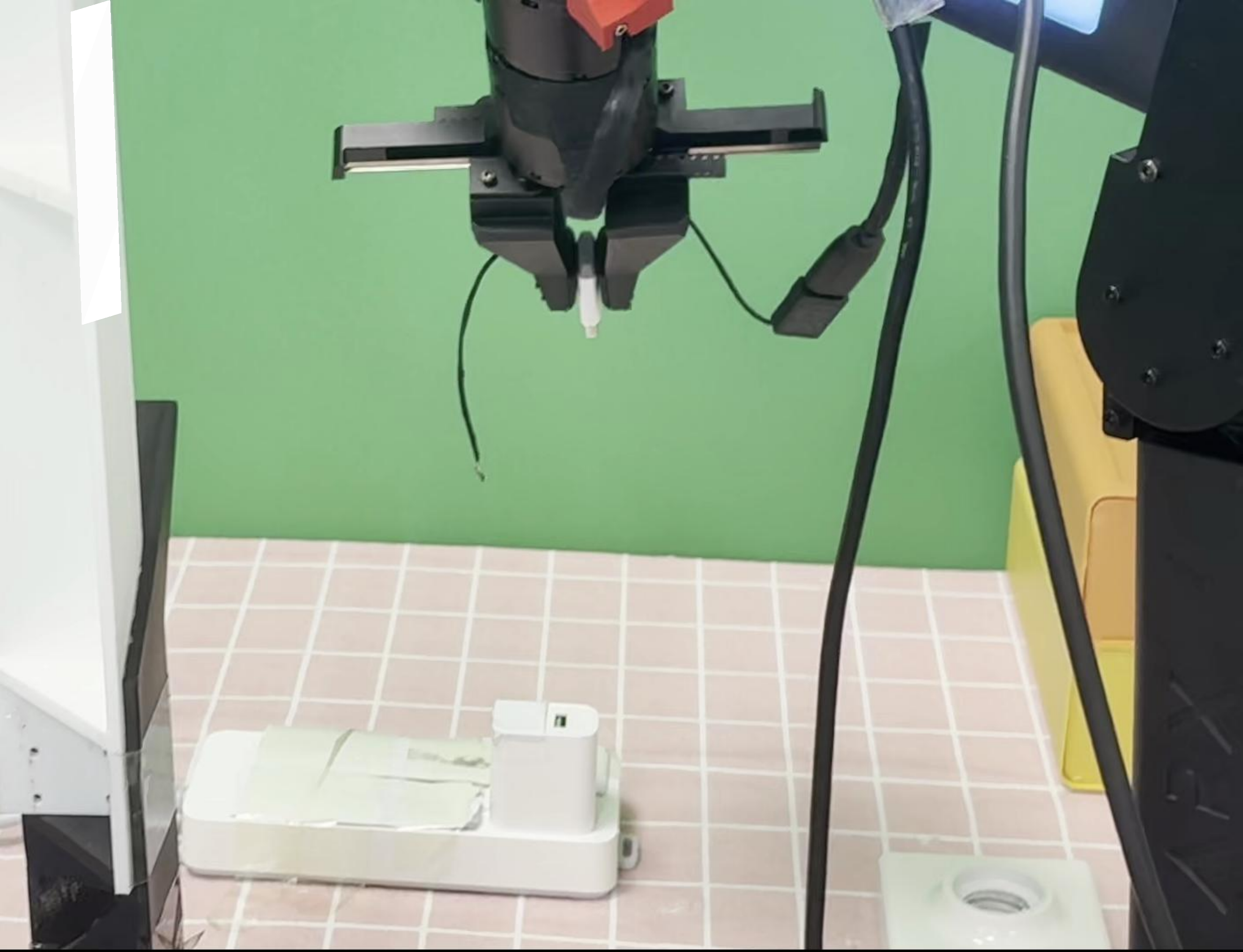}{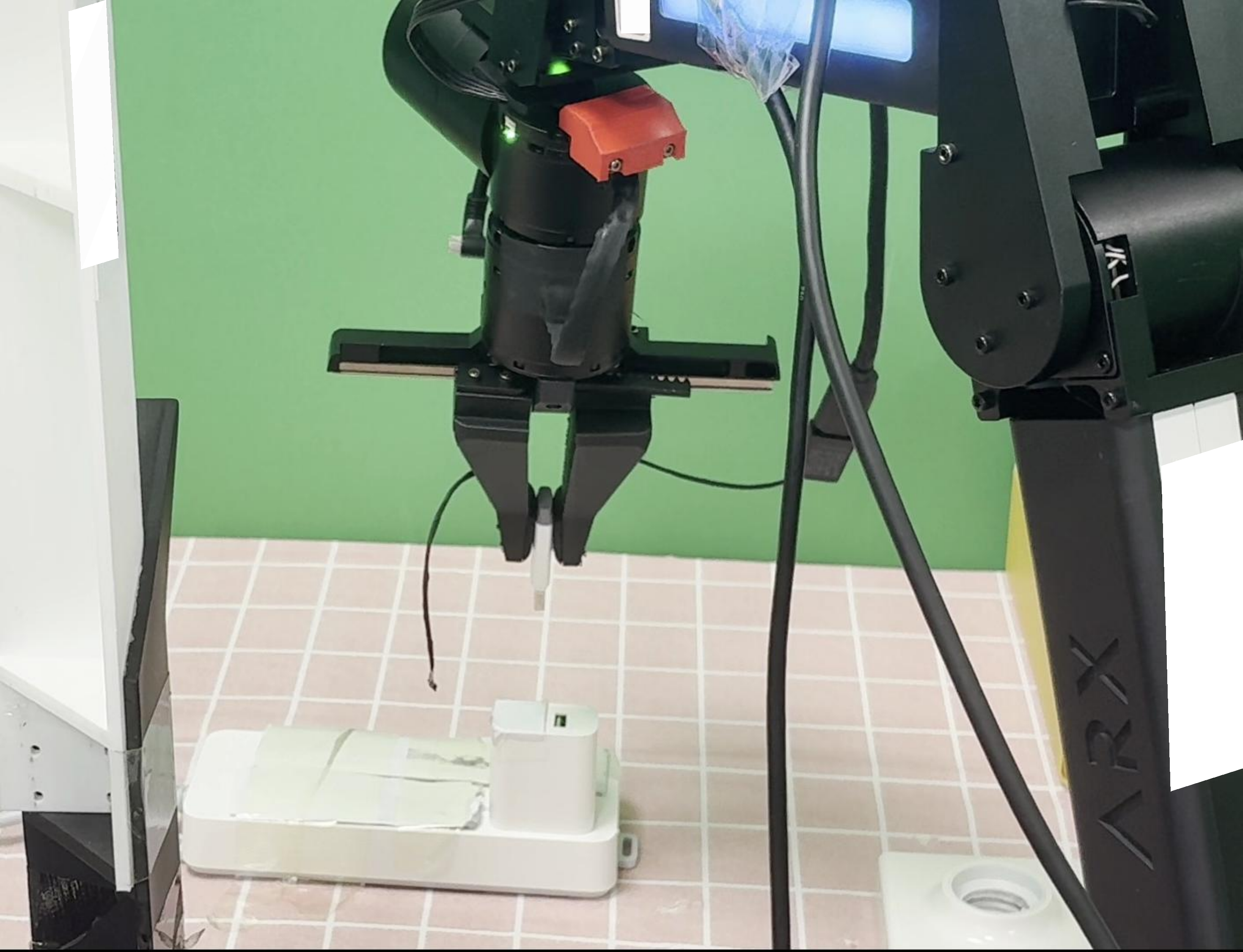}{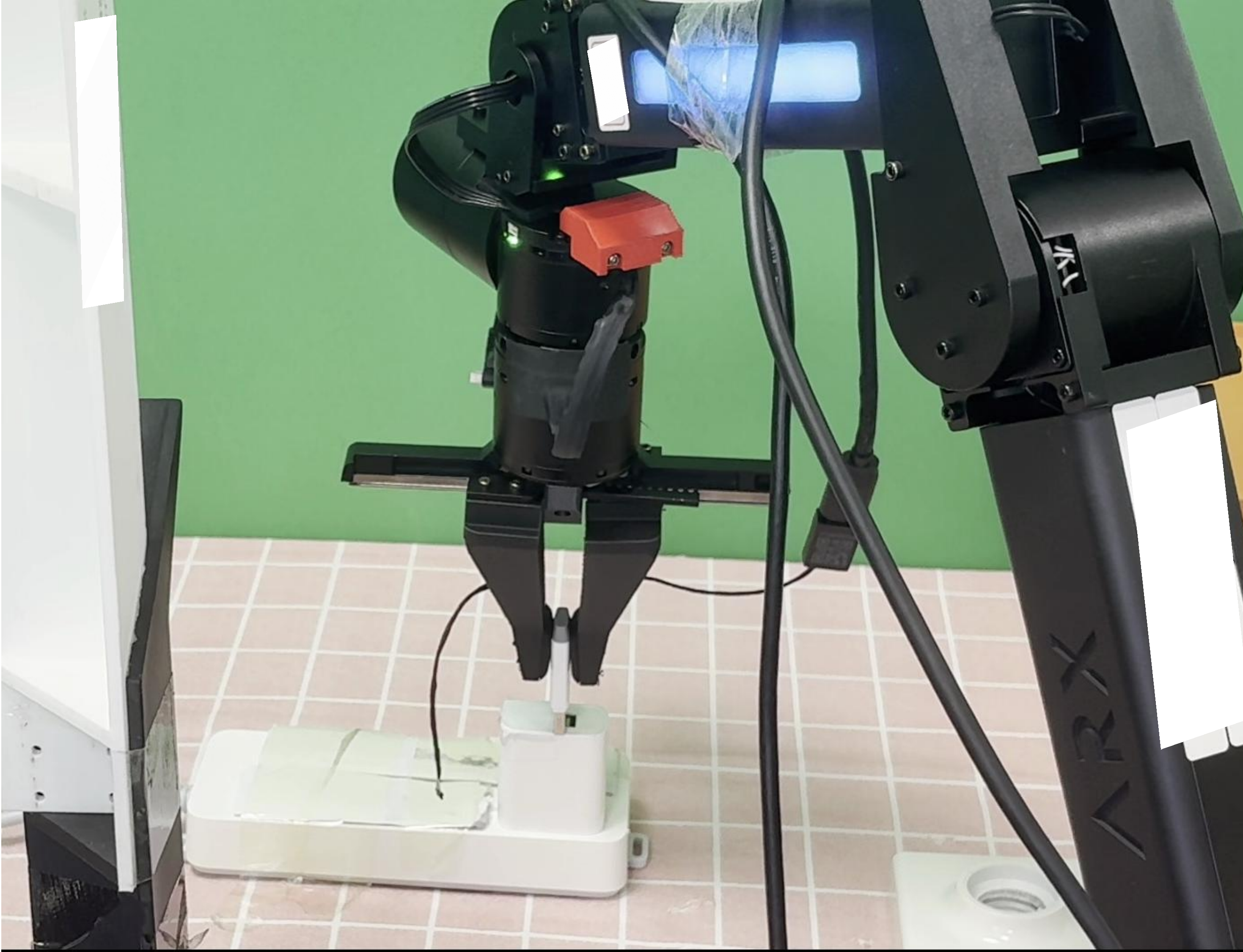}{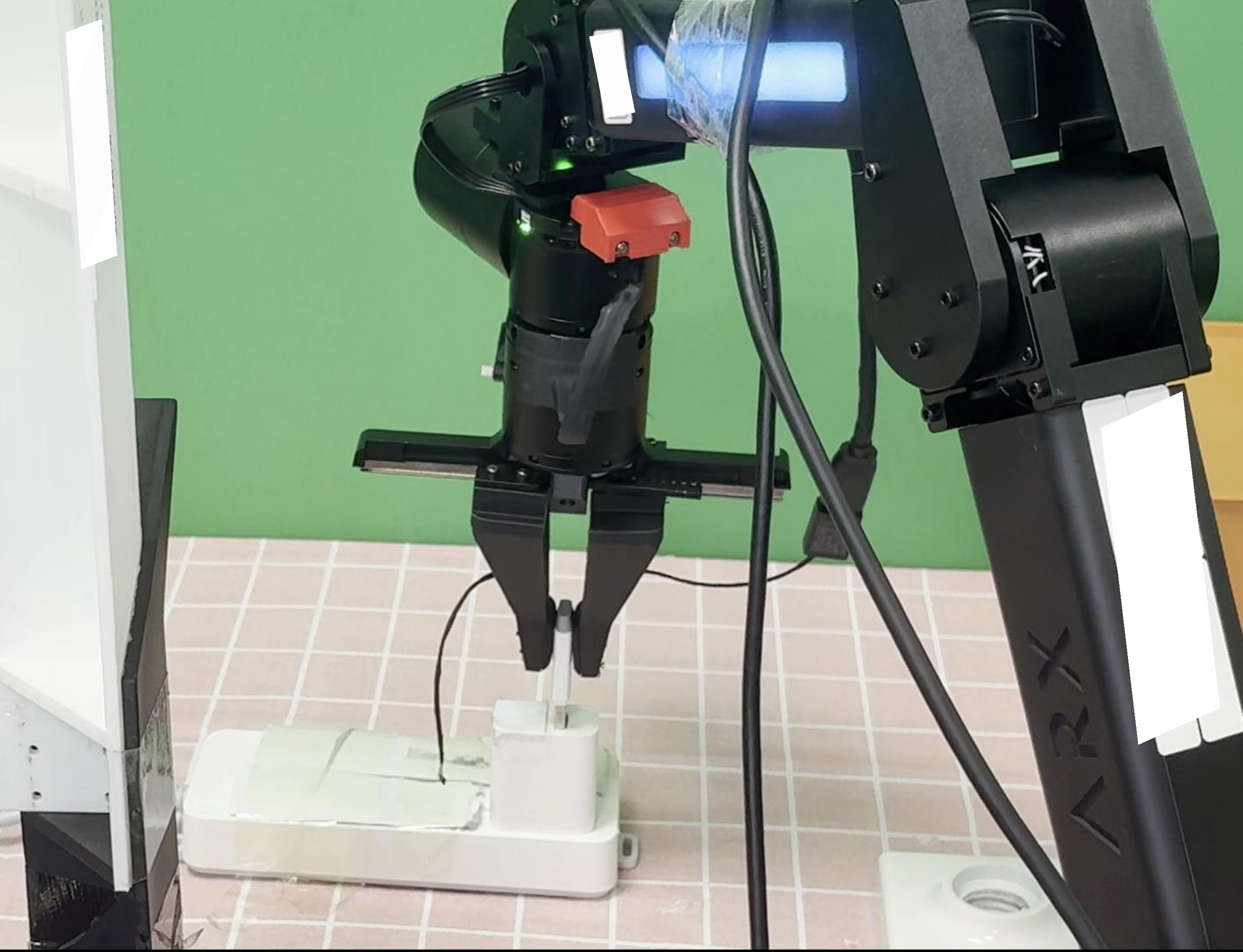}{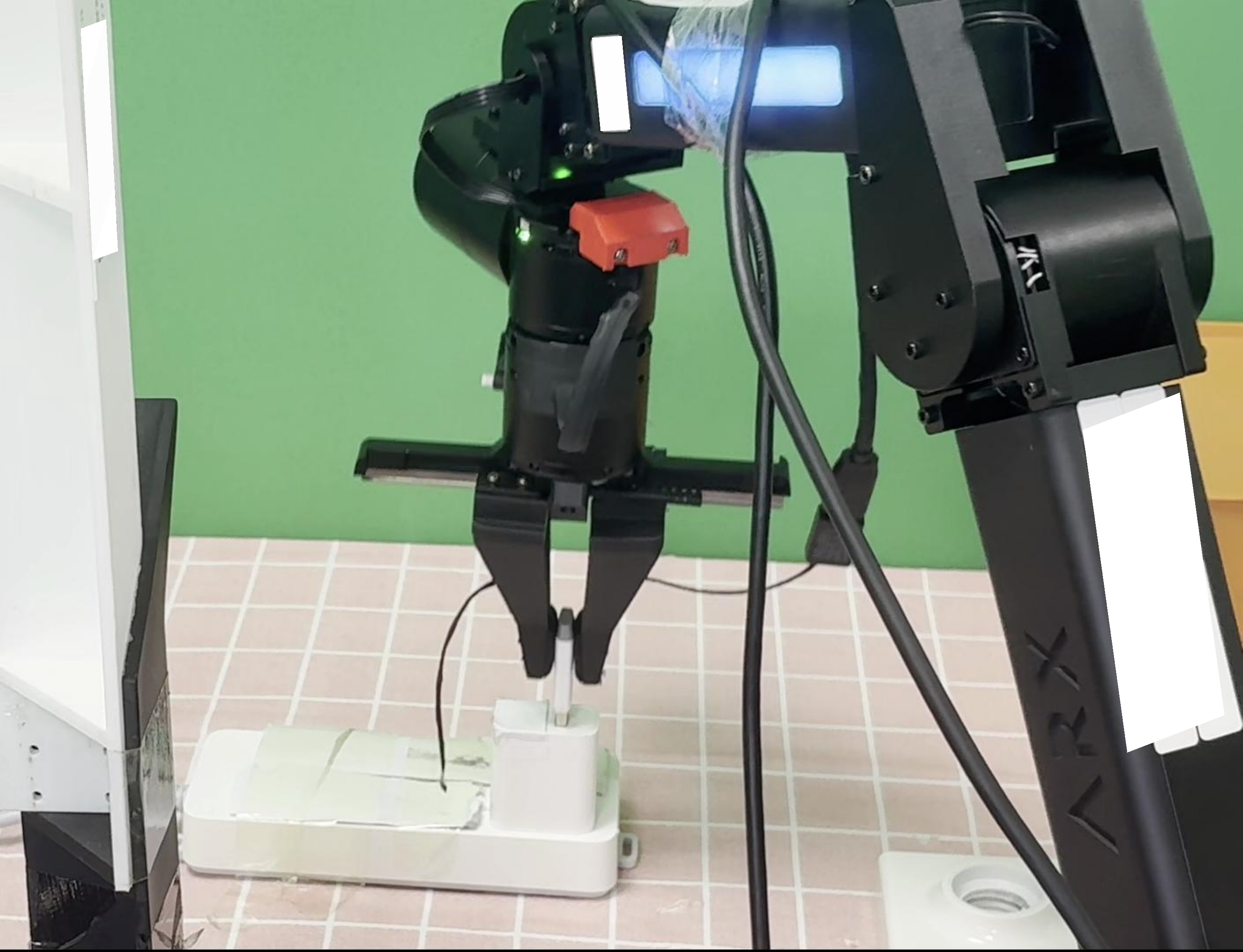}{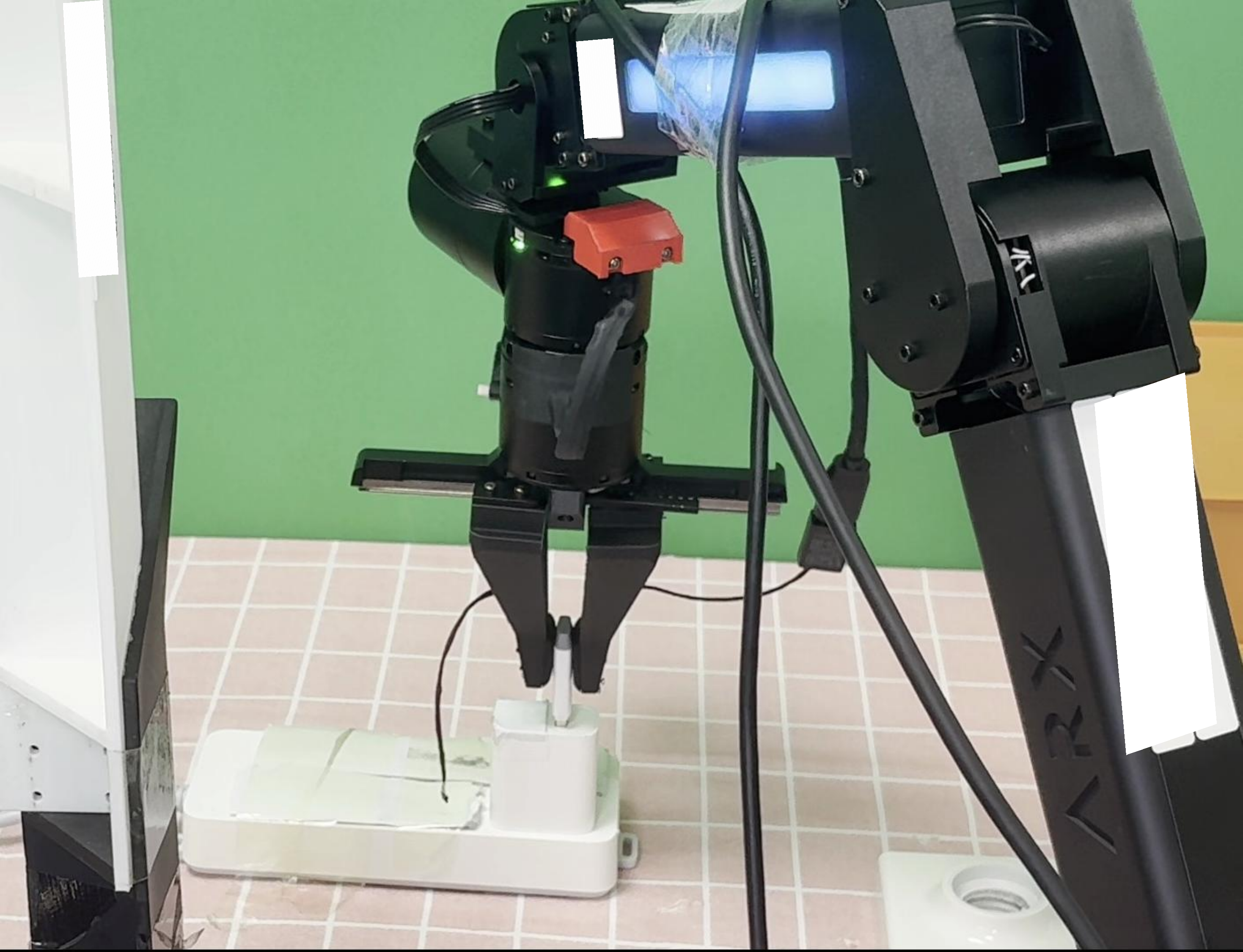}{USB Drive Insertion : Insert the USB drive into the slot.}

\vspace{0.24cm}
\apptaskrow{(c)}{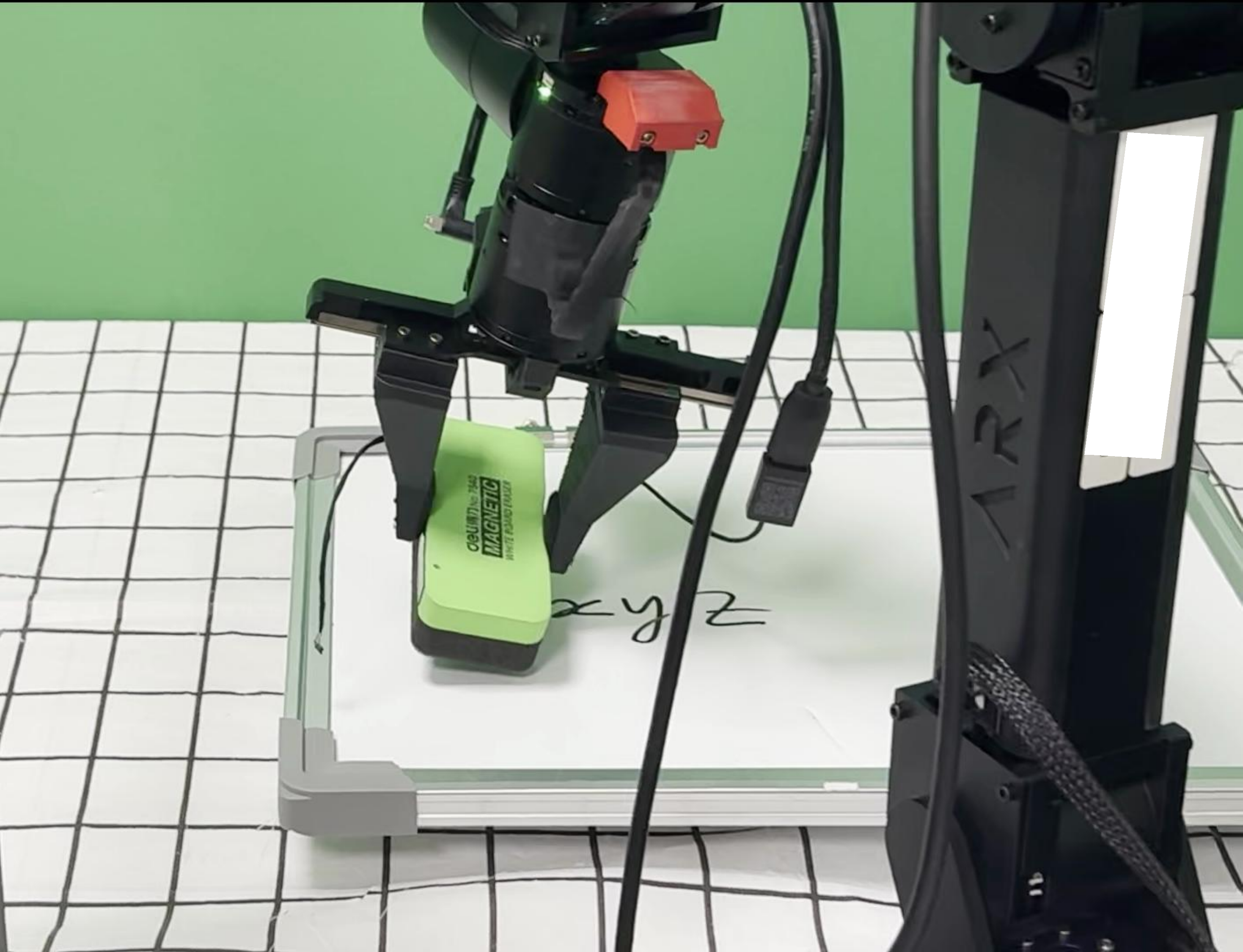}{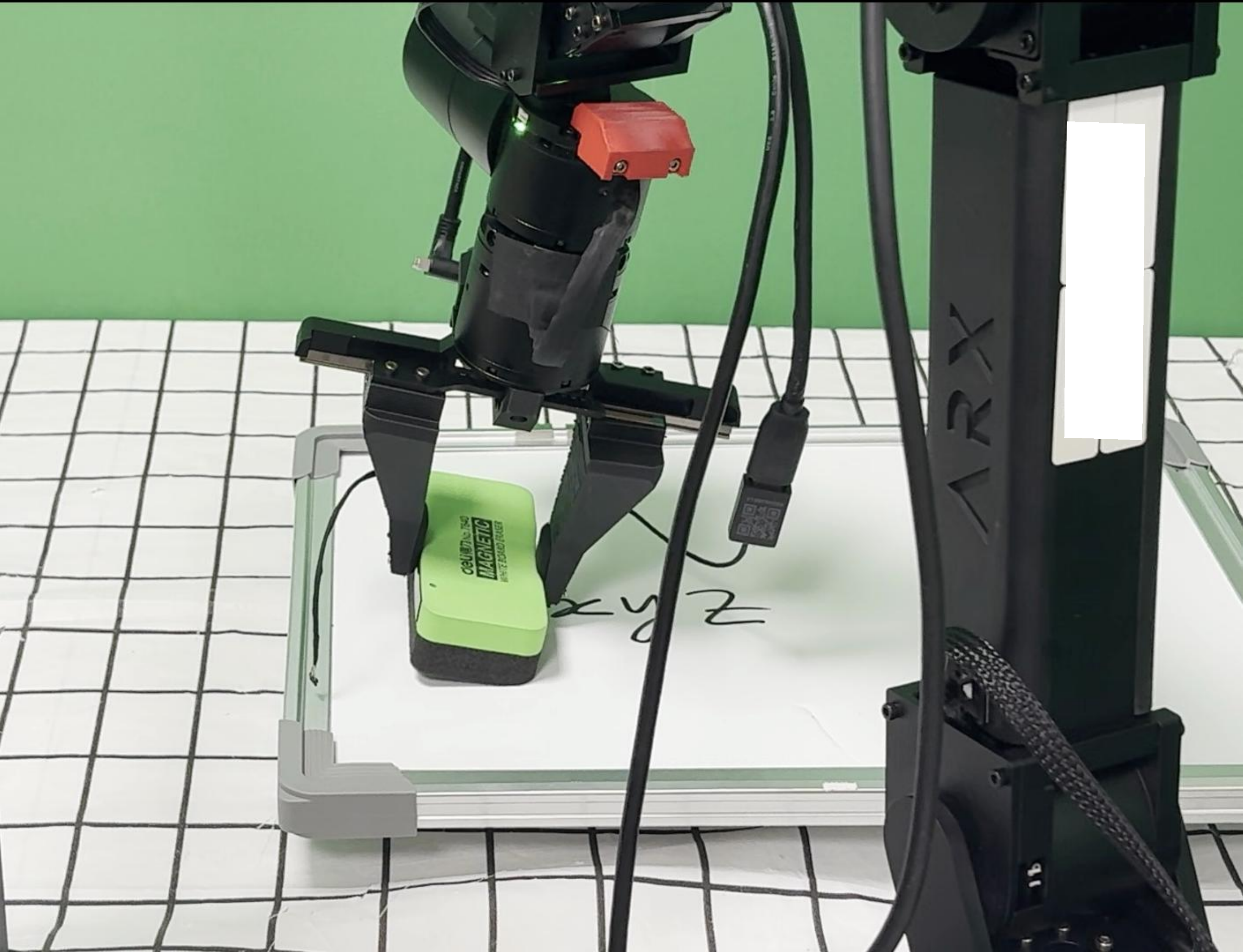}{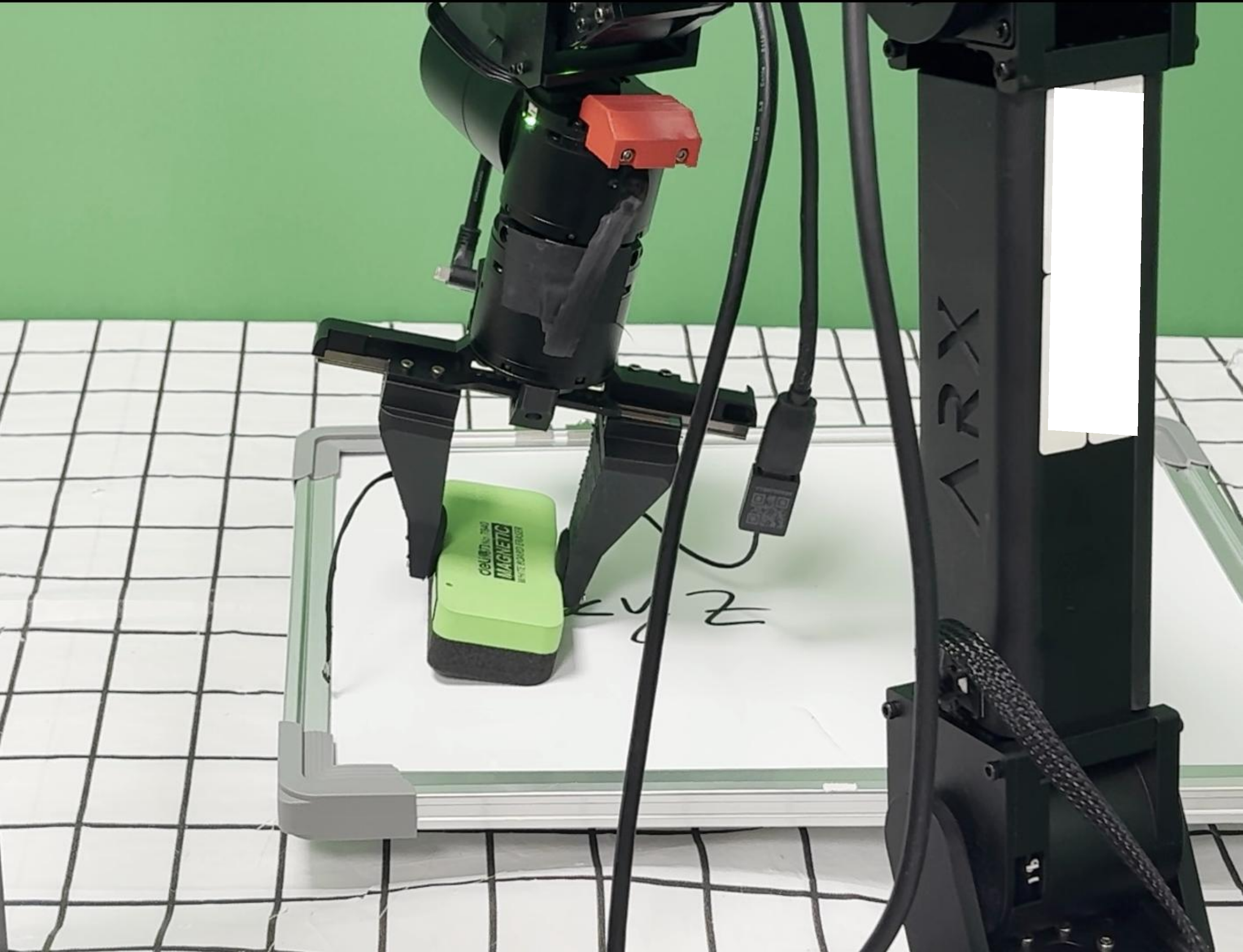}{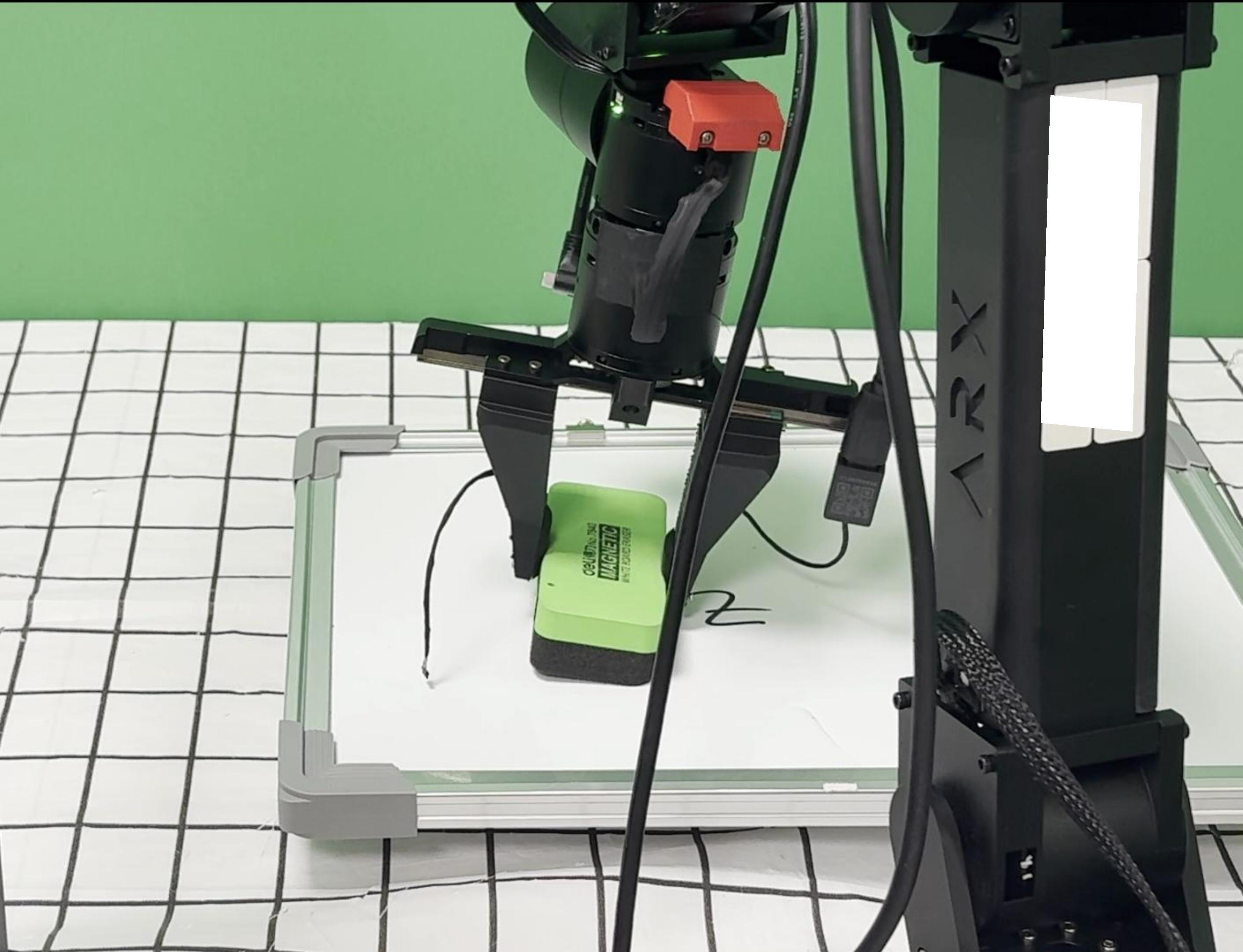}{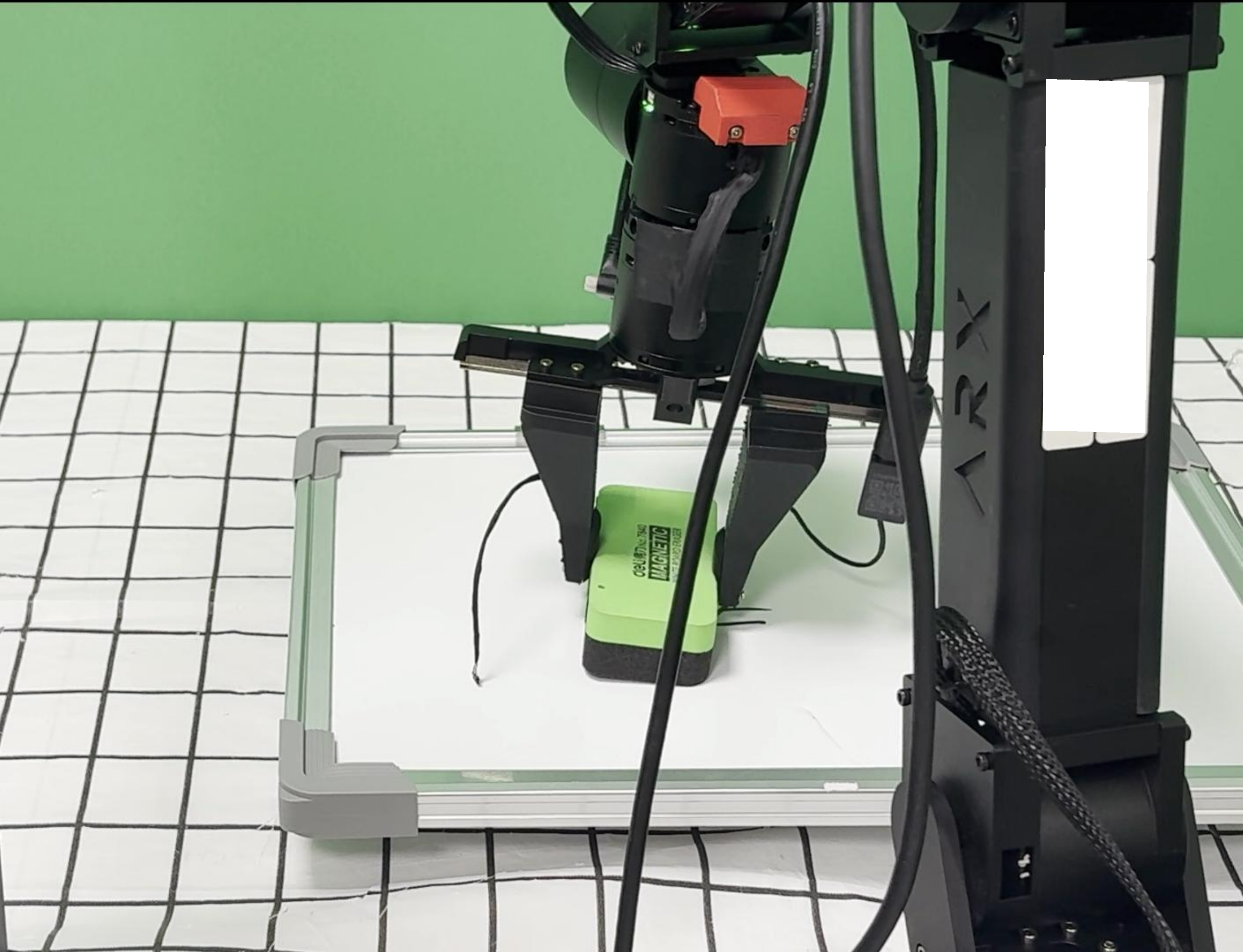}{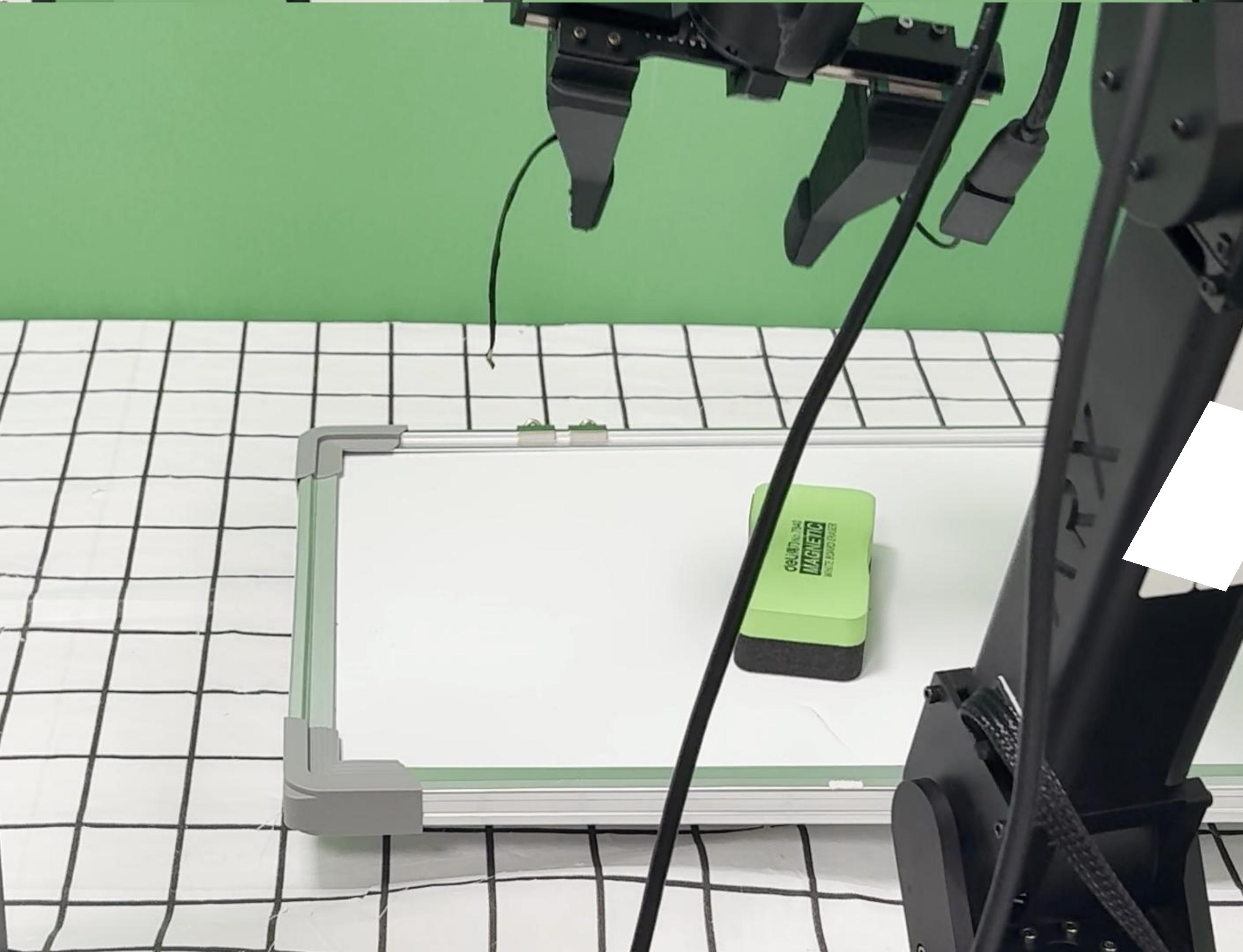}{Whiteboard Wiping : Erase the words on the whiteboard.}

\vspace{0.24cm}
\apptaskrow{(d)}{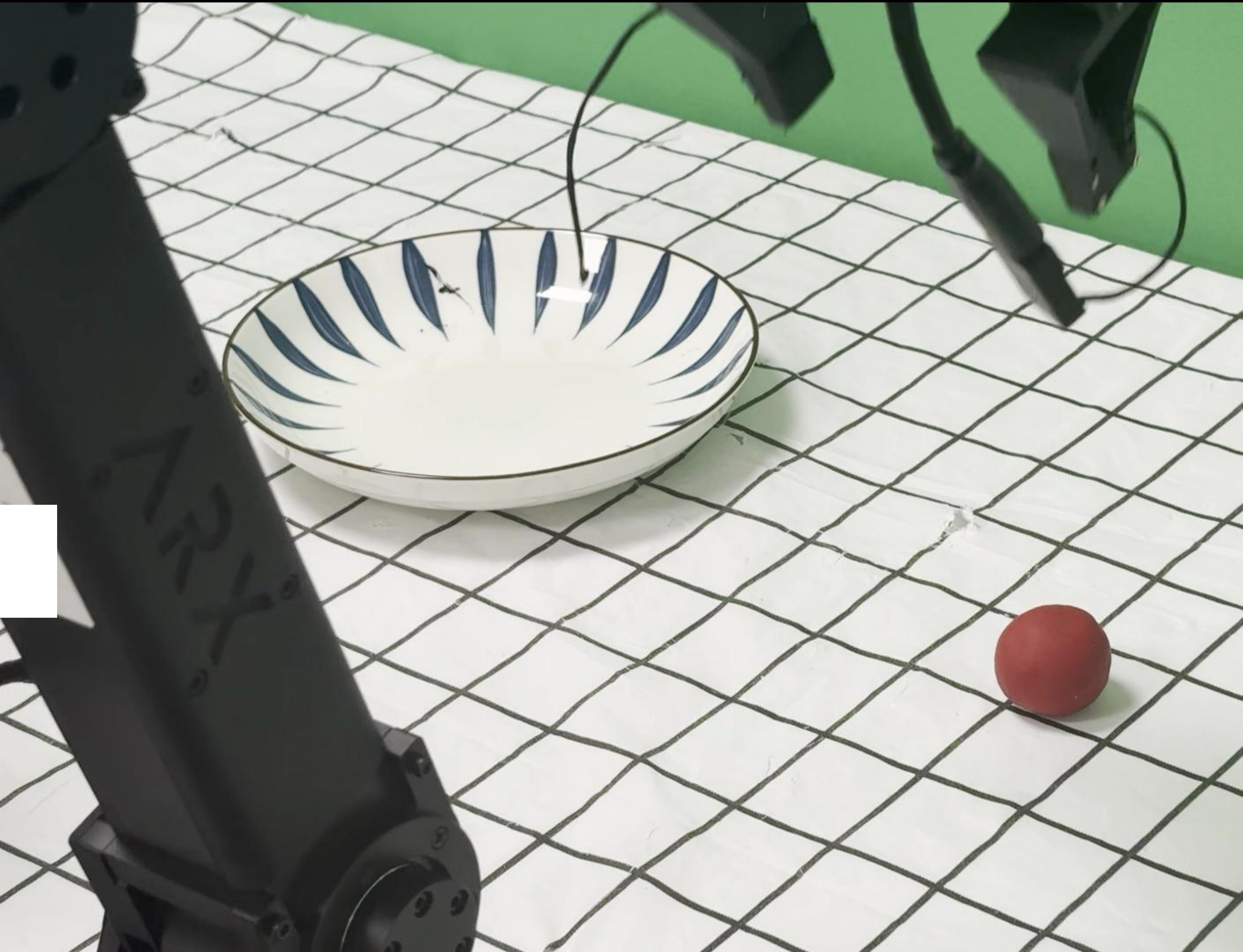}{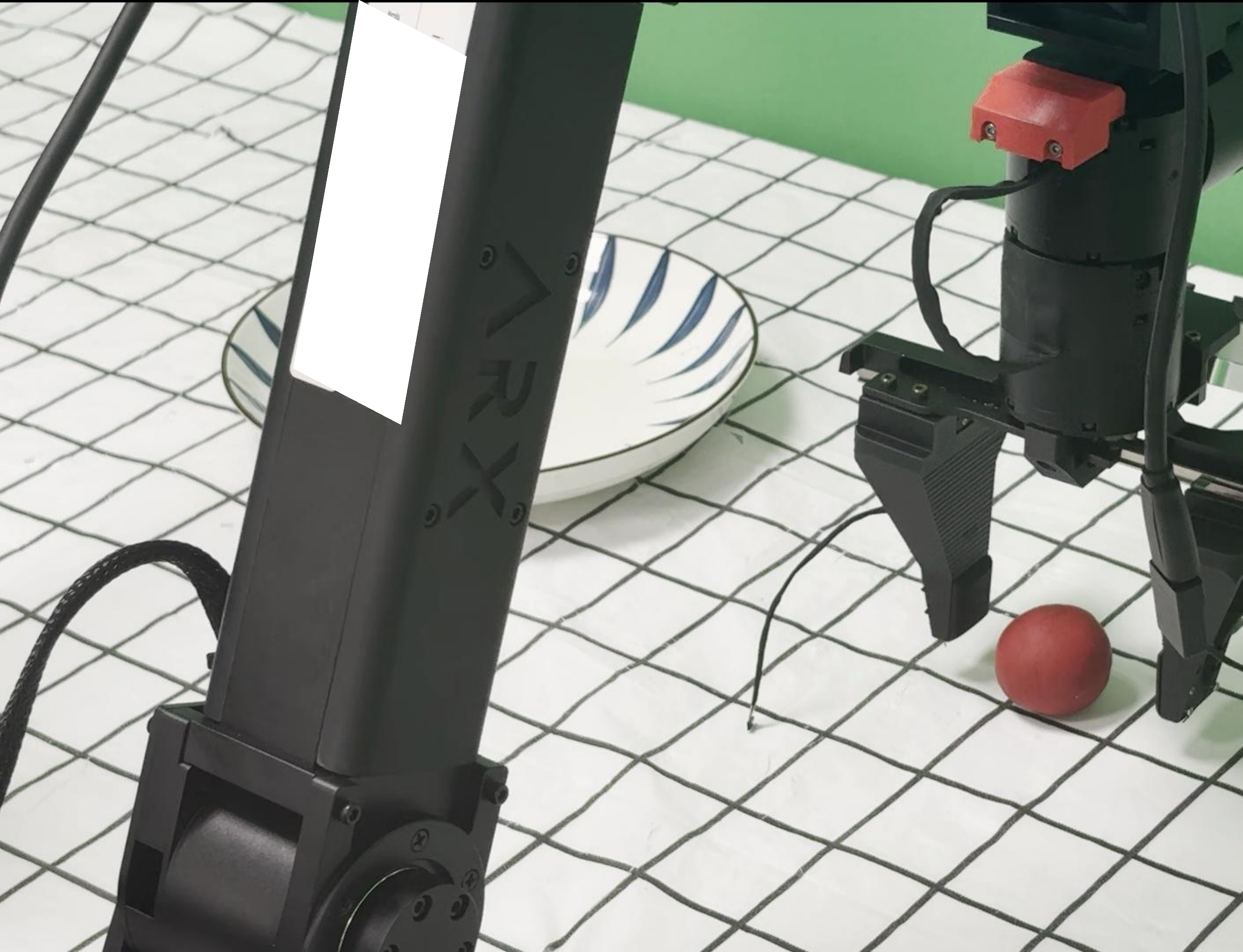}{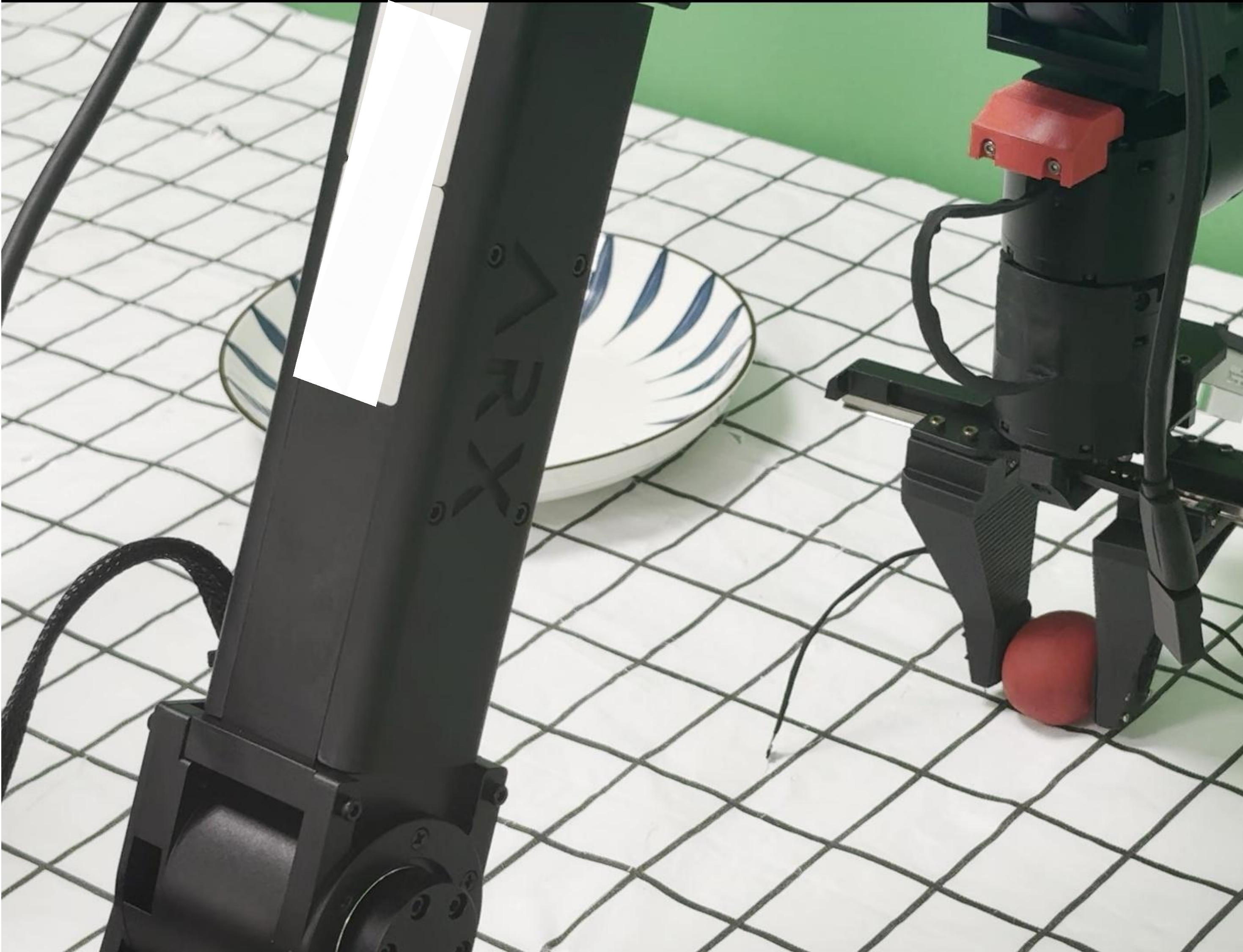}{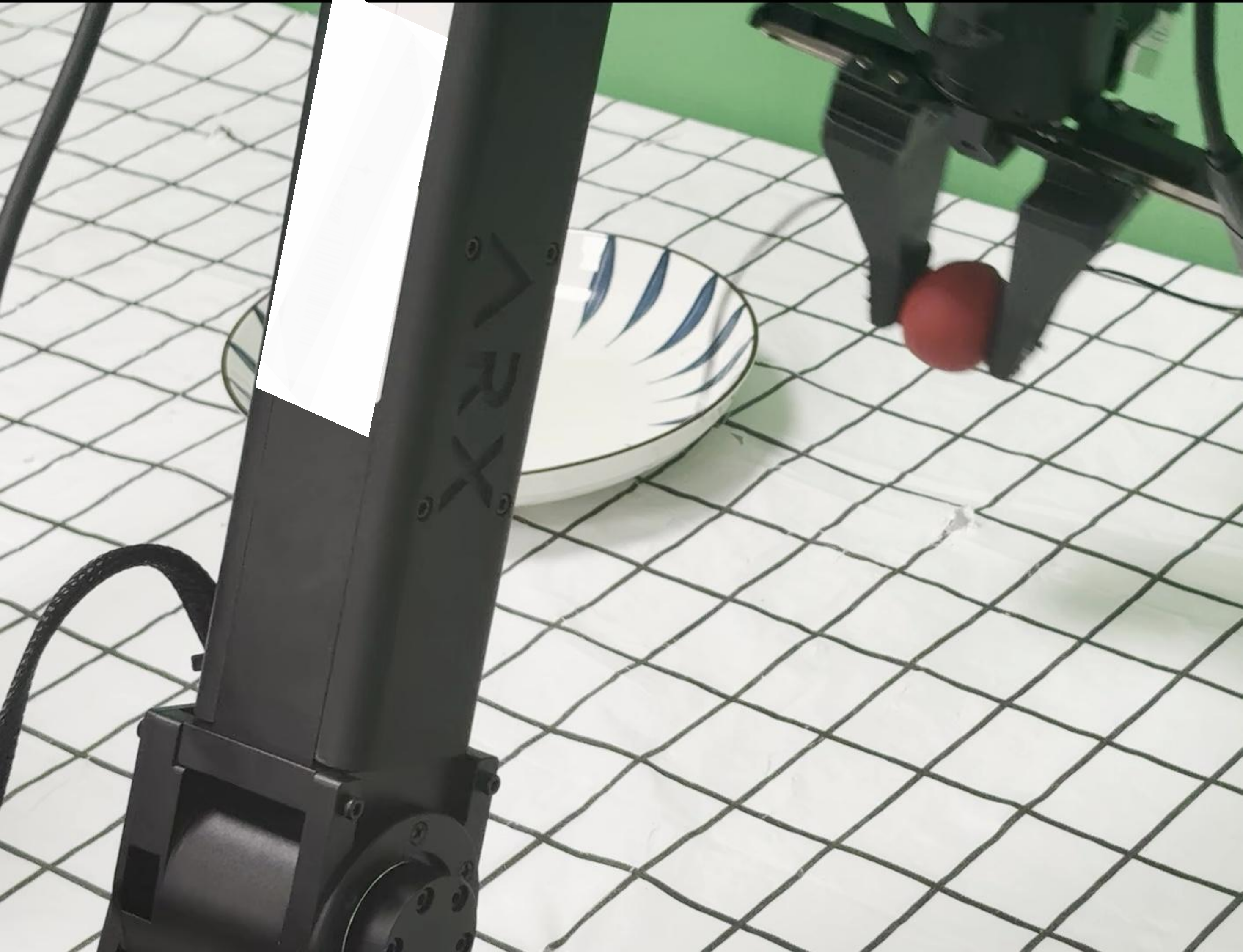}{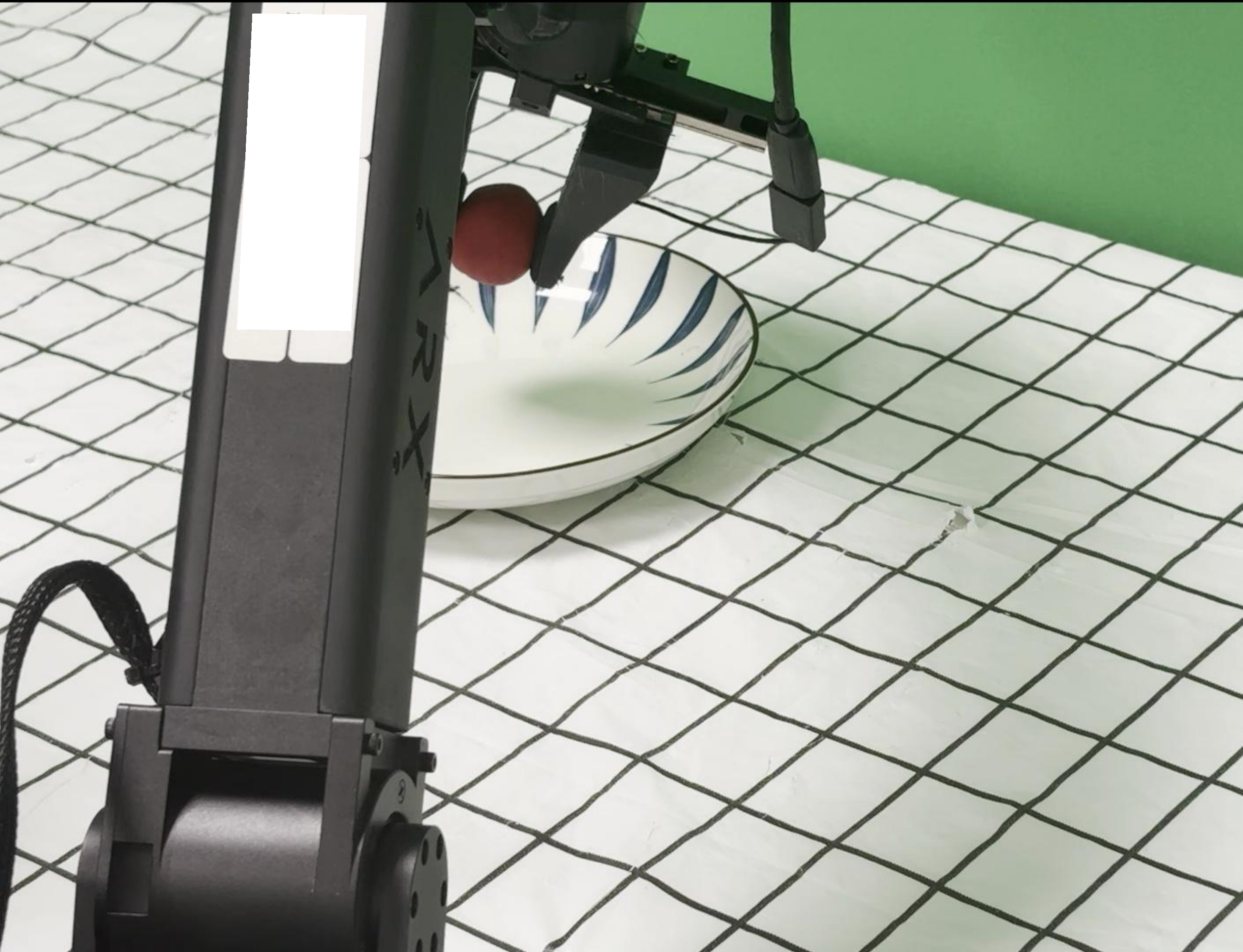}{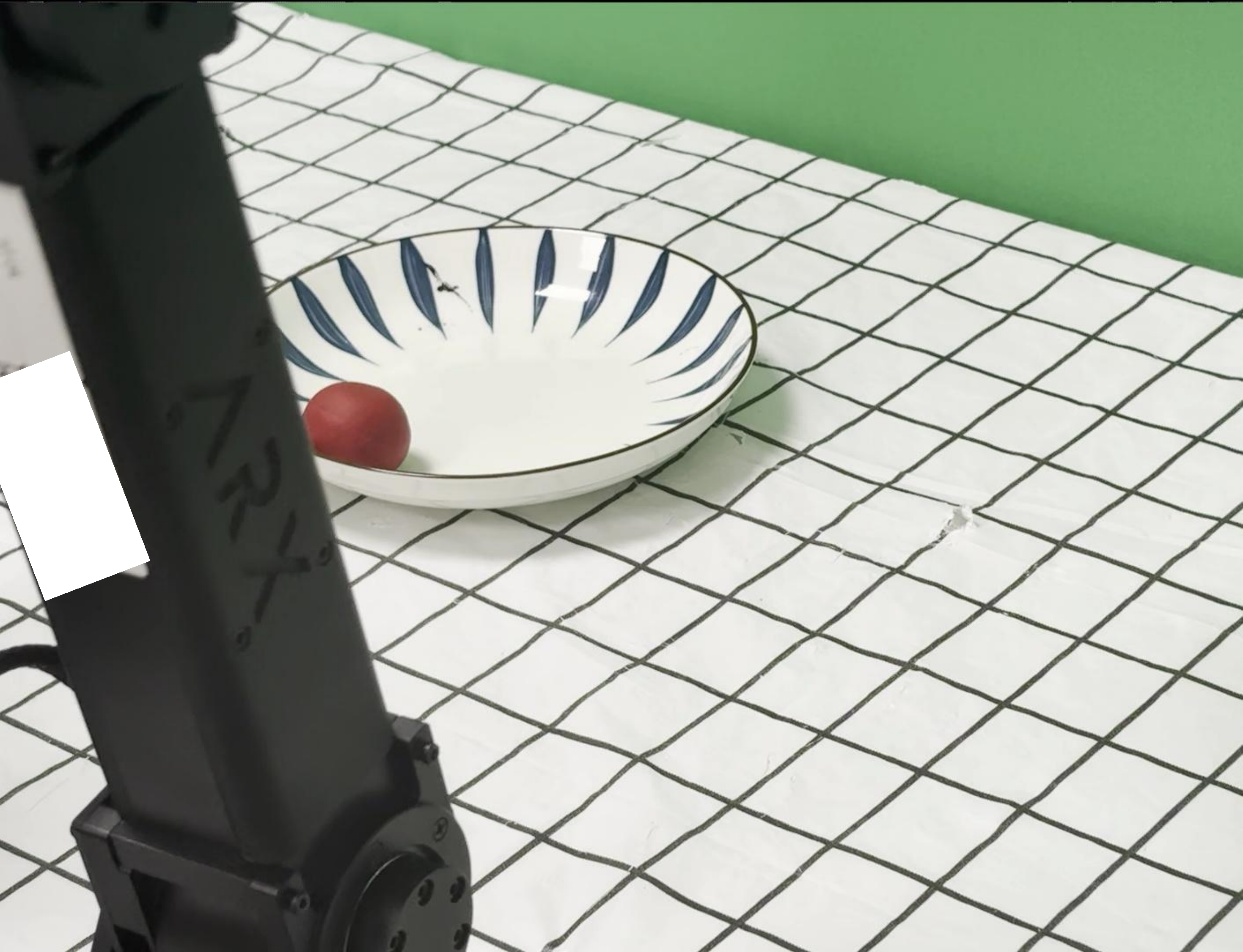}{Deformable Object Grasping : Gently clamp the clay ball onto the plate.}

\vspace{0.24cm}
\apptaskrow{(e)}{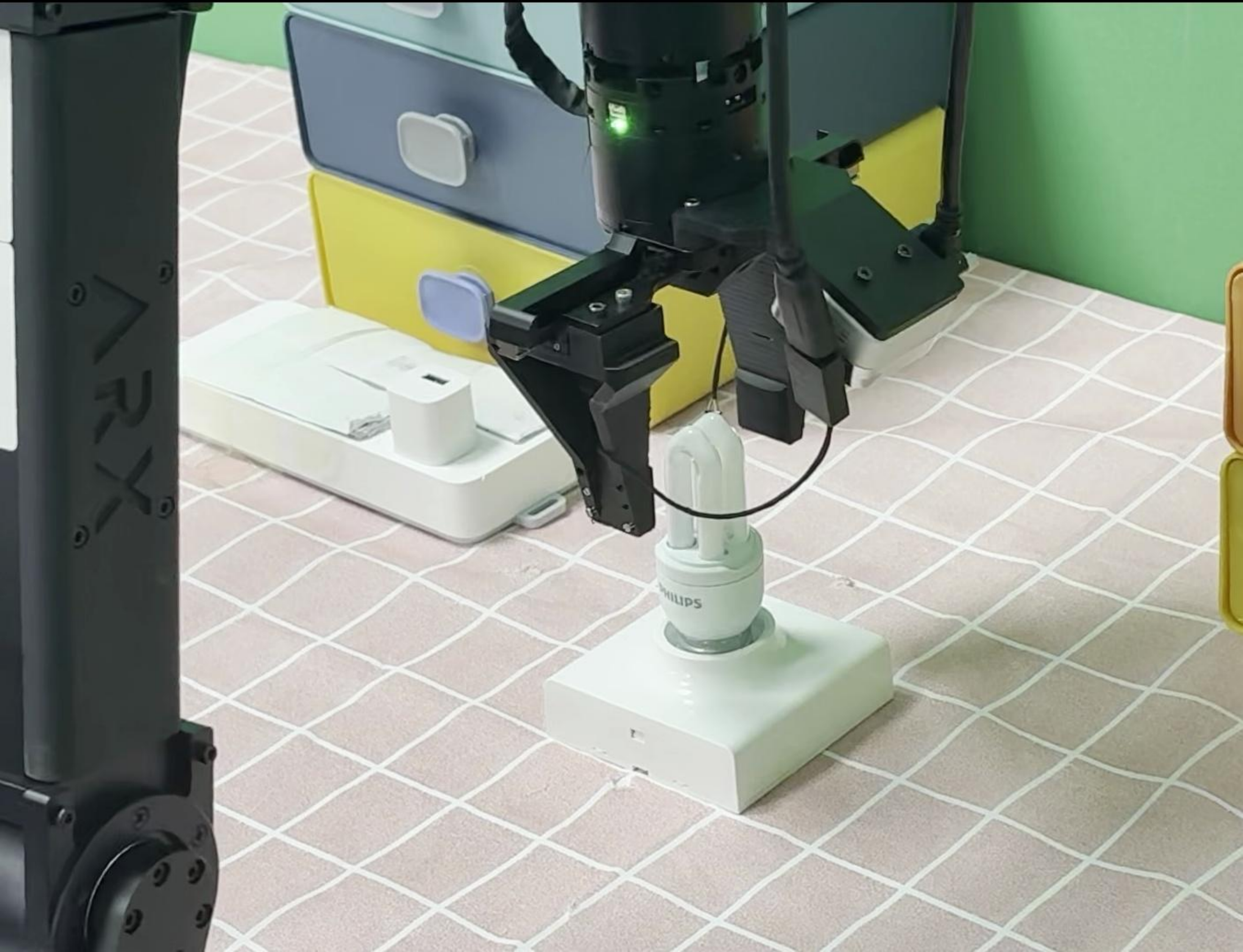}{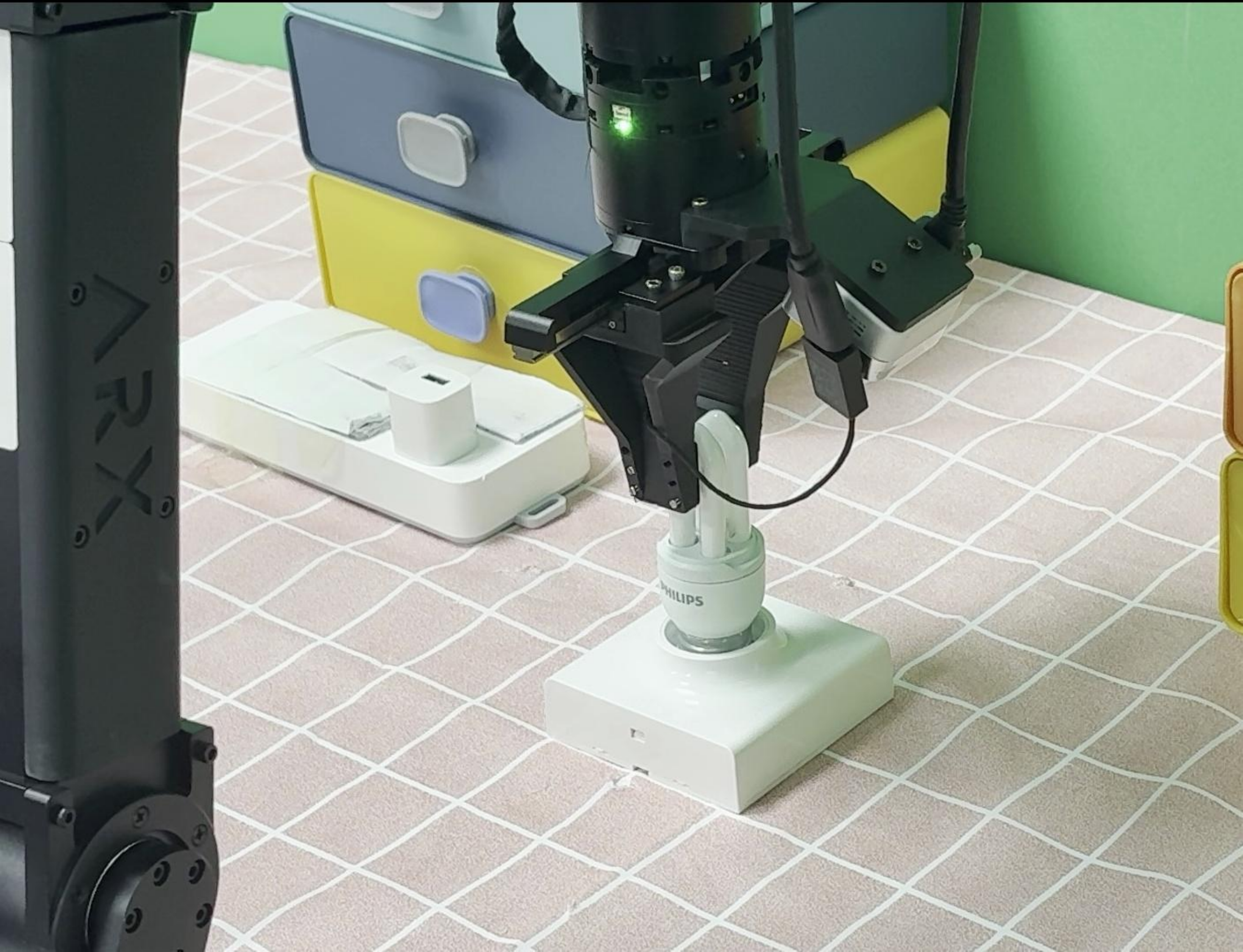}{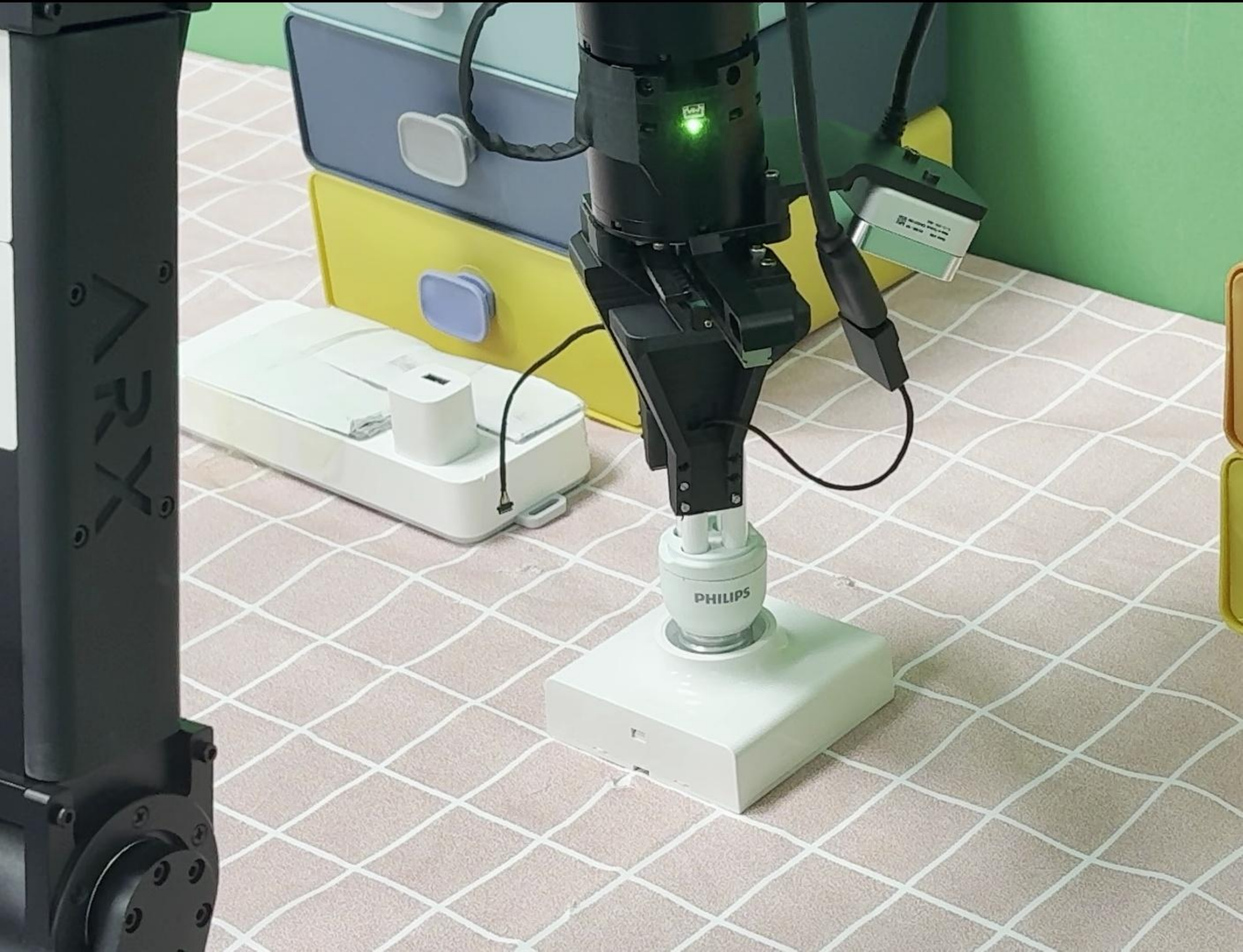}{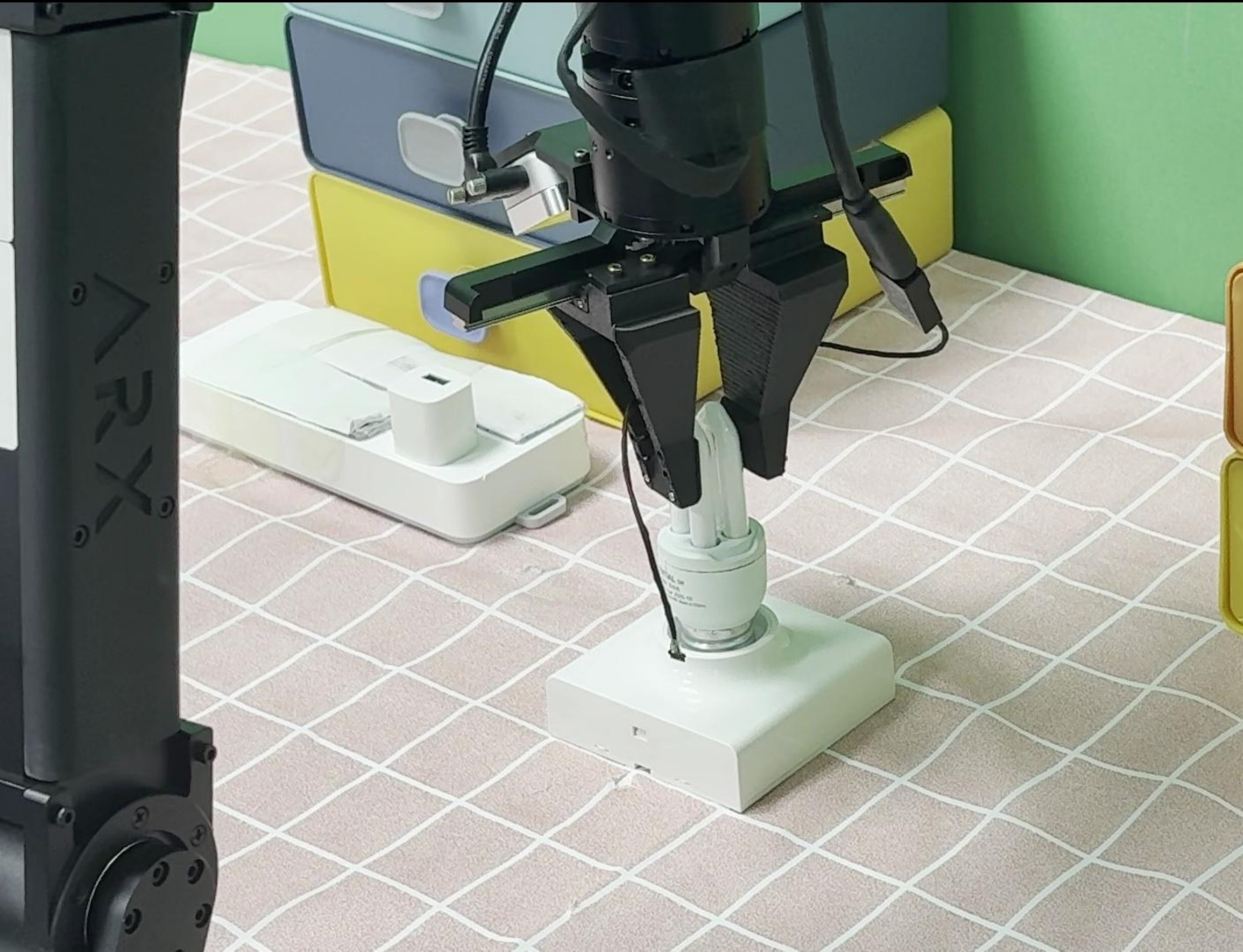}{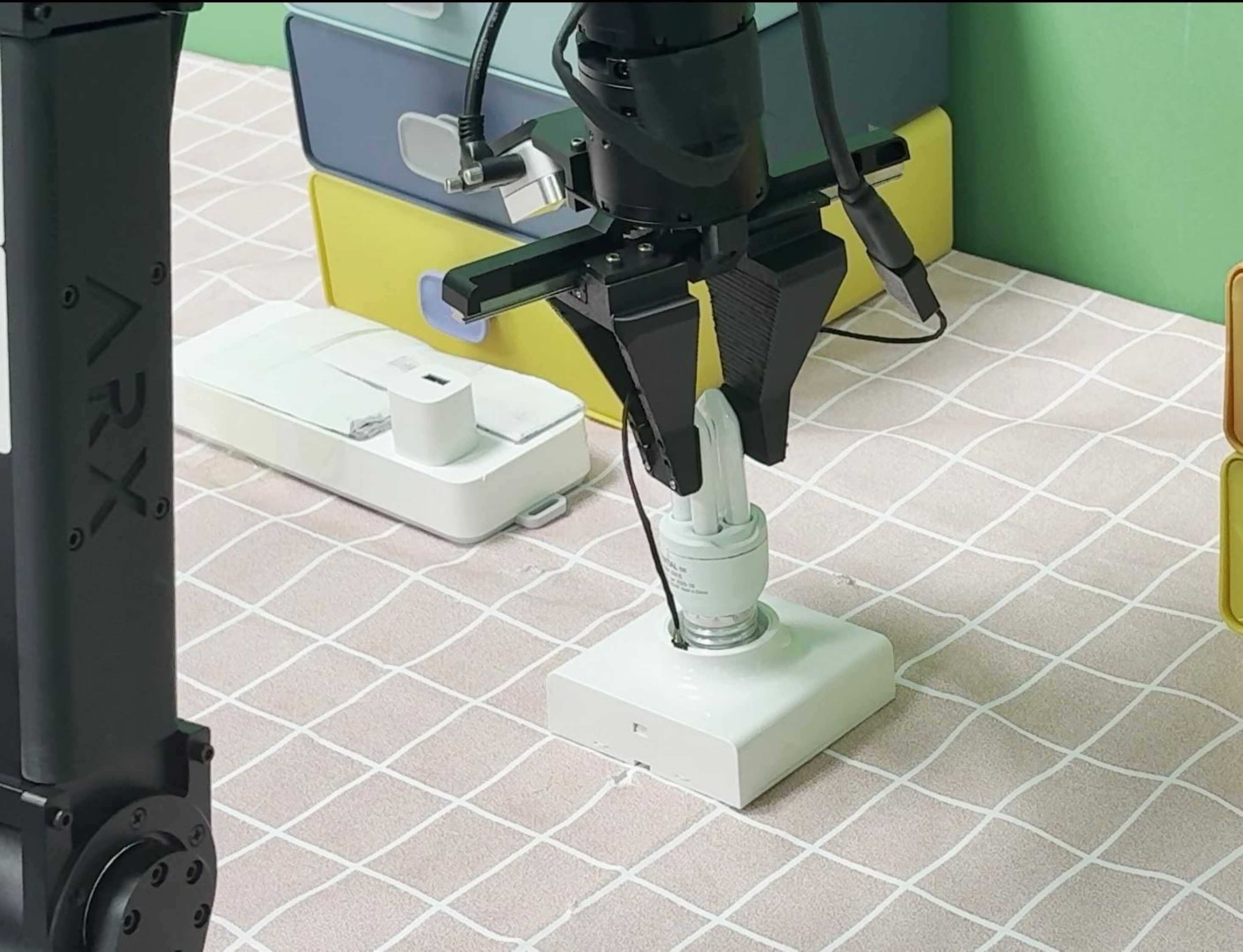}{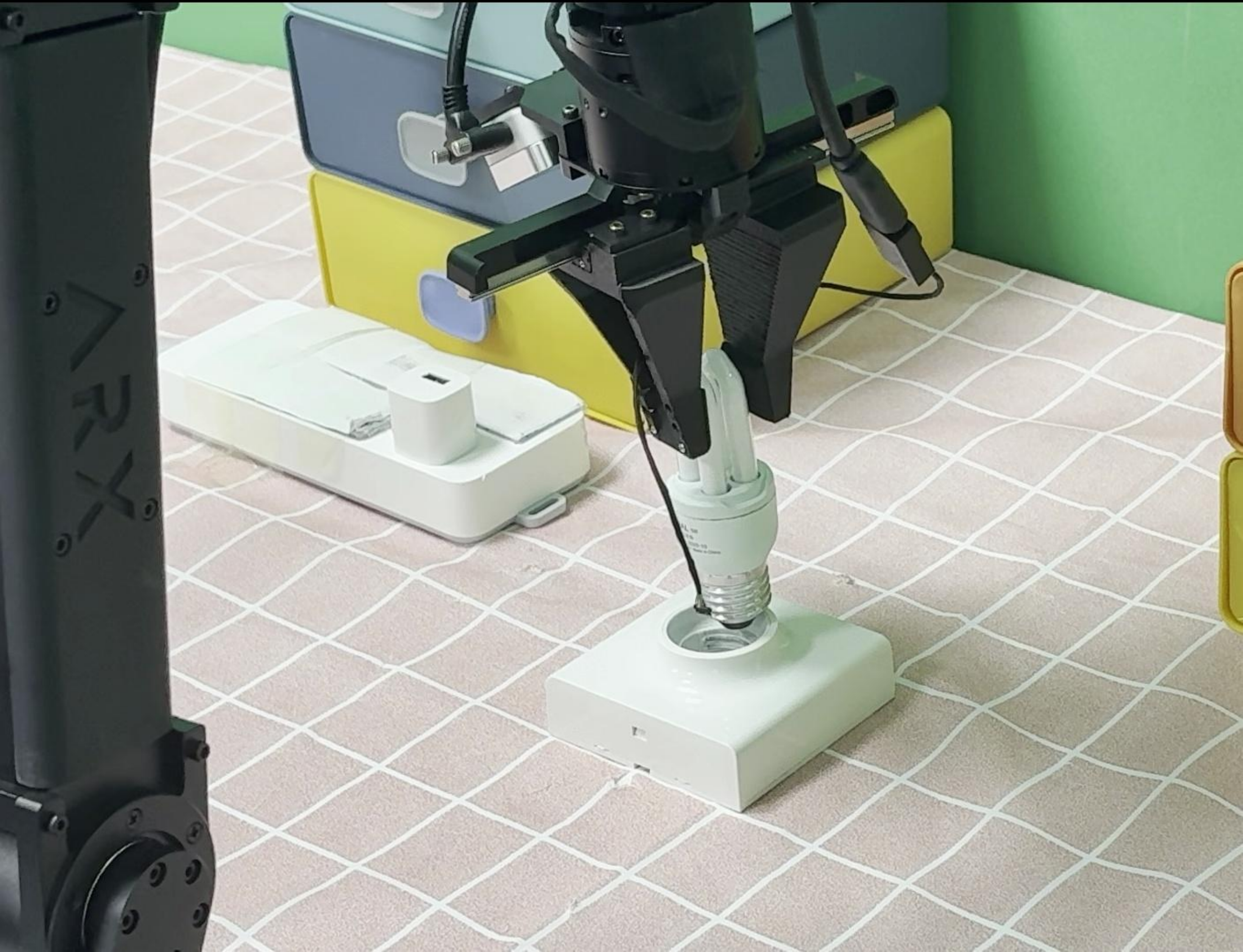}{Light Bulb Unscrewing : Pull the bulb out of the base.}

\vspace{0.24cm}
\caption{\textbf{Qualitative examples of successful real-world rollouts.} Uniformly sampled frames from successful rollouts of Plug Insertion, USB Drive Insertion, Whiteboard Wiping, Deformable Object Grasping, and Light Bulb Unscrewing are arranged chronologically to illustrate the execution process.}
\label{fig:app_qualitative_all_tasks}
\end{figure}

\end{document}